\documentclass[lettersize,journal]{IEEEtran}
\usepackage{amsmath,amsfonts}
\usepackage{algorithmic}
\usepackage{algorithm}
\usepackage{array}
\usepackage[caption=false,font=normalsize,labelfont=sf,textfont=sf]{subfig}
\usepackage{textcomp}
\usepackage{stfloats}
\usepackage{url}
\usepackage{verbatim}
\usepackage{graphicx}
\usepackage{cite}
\usepackage{tabularx}
\usepackage{booktabs}
\usepackage{multirow}
\usepackage{amssymb}
\usepackage{mathrsfs}
\usepackage{makecell}
\usepackage{booktabs}
\usepackage{orcidlink}
\usepackage[labelfont=bf,format=plain,justification=raggedright,singlelinecheck=false]{caption}
\begin{document}

\title{Scalable Hierarchical Reinforcement Learning for Hyper Scale Multi-Robot Task Planning}

\author{Xuan Zhou, Xiang Shi, Lele Zhang,
	Chen Chen\IEEEmembership{~Member,~IEEE},
	 Hongbo Li, Lin Ma, \\
	  Fang Deng\IEEEmembership{~Senior~Member,~IEEE}, and Jie Chen\IEEEmembership{~Fellow,~IEEE}
	\thanks{Xuan Zhou, Lele Zhang, Chen Chen and Fang Deng are with the State Key Lab of Autonomous Intelligent Unmanned Systems, School of Automation, Beijing Institute of Technology, Beijing 100081, China (e-mail:zhouxuanbit@163.com,  zhanglele@bit.edu.cn, xiaofan@bit.edu.cn, dengfang@bit.edu.cn).}	 
	\thanks{Xiang Shi is with the Department of Automation, Tsinghua University, Beijing 10084, China (email:shi-xiang@tsinghua.edu.cn).}
	\thanks{Hongbo Li is with the Beijing Geek+ Technology Co., Ltd, Beijing 100000, China (e-mail: jason.li@geekplus.com).} 
	\thanks{Lin Ma is with Zhejiang Cainiao Supply Chain
		Management Company Ltd., Hangzhou 311101, China (e-mail: mark.ml@
		cainiao.com).}	
	\thanks{Jie Chen is with the State Key Lab of Autonomous Intelligent Unmanned Systems, Beijing Institute of Technology, Beijing
		100081, China, and also with Shanghai Research Institute for Intelligent
		Autonomous Systems, Tongji University, Shanghai 200092, China (e-mail:
		chenjie@bit.edu.cn).}
}

\maketitle

\begin{abstract}
	To improve the efficiency of warehousing system and meet huge customer orders, we aim to solve the challenges of dimension disaster and dynamic properties in hyper scale multi-robot task planning (MRTP) for robotic mobile fulfillment system (RMFS).
	Existing research indicates that hierarchical reinforcement learning (HRL) is an effective method to reduce these challenges. 
	Based on that, we construct an efficient multi-stage HRL-based multi-robot task planner for hyper scale MRTP in RMFS, and
	the planning process is represented with a special temporal graph topology.
	Following its temporal logic, only critical events deserve attention, so system can sample efficiently when training and hold dynamic response in execution.  
	To ensure optimality, the planner is designed with a centralized architecture, but it also brings the challenges of scaling up and generalization that require policies to maintain performance for various unlearned scales and maps. To tackle these difficulties, we first construct a hierarchical temporal attention network (HTAN) to ensure basic ability of handling inputs with unfixed lengths, and then design multi-stage curricula for hierarchical policy learning to further improve the scaling up and generalization ability while avoiding catastrophic forgetting.	
	Additionally, we notice that policies with hierarchical structure suffer from unfair credit assignment that is similar to that in multi-agent reinforcement learning, inspired of which, we propose a hierarchical reinforcement learning algorithm with counterfactual rollout baseline to improve learning performance.
	Experimental results demonstrate that our planner outperform other state-of-the-art methods on various MRTP instances  in both simulated and real-world RMFS. Also, our planner can successfully scale up to hyper scale MRTP instances in RMFS with up to 200 robots and 1000 retrieval racks 
	on unlearned maps while keeping superior performance over other methods.  
\end{abstract}


\section{Introduction}
\label{sec:introduction}

\IEEEPARstart{T}ASK planning has long been a significant area of research in multi-robot systems, exerting a crucial influence on the intelligent capabilities and operational efficiency of the cluster system \cite{trends}. 
The process of multiple robot task planning (MRTP) focuses on the determination of scheme both in task scheduling (TS), task allocation (TA) and task decomposition (TD) to accomplish a particular objective \cite{MRTP}.
%\IEEEpubidadjcol
Specifically, TS aims to determine the sequence of tasks assigned to each robot \cite{schedule}. TA involves assigning specific tasks to either an individual robot or a group of robots \cite{allocation}. TD focuses on solving complex tasks by breaking down the overall task into simpler and more manageable sub-tasks \cite{decompose}.
\begin{figure}[!t]
	\centering
	\includegraphics[width=3.4in]{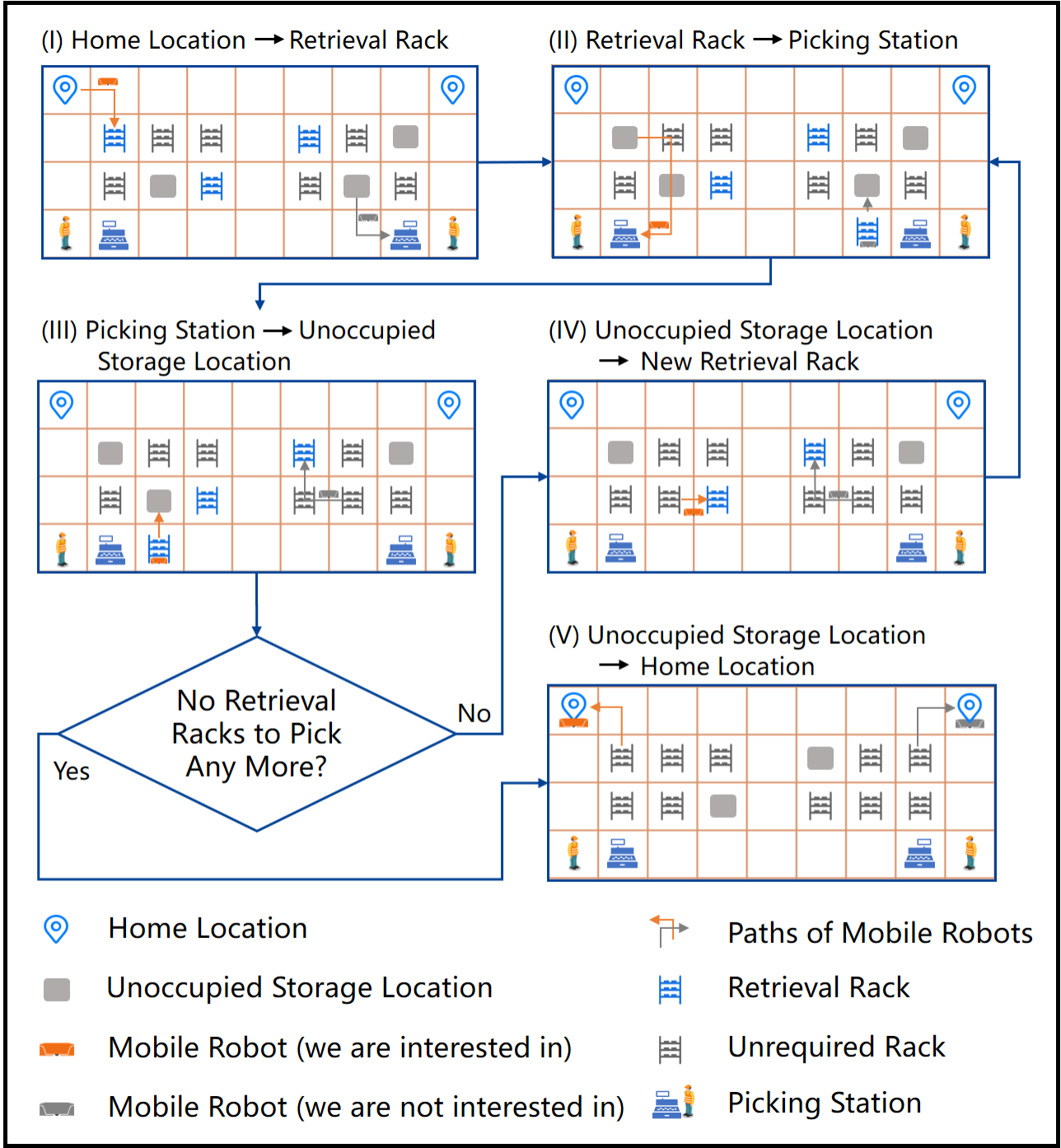}
	\caption{The operational process for MRTP in RMFS.}
	\label{problem}
\end{figure}

In this article, we focus on a specific MRTP problem abstracting from the robot mobile fulfillment system (RMFS) \cite{RMFS}, which is a typical multiple robot system built to enhance warehouse operational efficiency \cite{Warehouse}.
In RMFS, the retrieval racks required by the top-level order will be delivered by mobile robots to their required picking stations, in which case, pickers don't need to walk to the rack locations to pick up cargoes any more, and this will greatly save labor cost and improve the picking efficiency.
A description of the operational process for RMFS is shown in Fig. \ref{problem}, where multiple mobile robots, movable racks, unoccupied storage locations, and picking stations are scattered in layout and they constitute the elements of this system. Initially, the mobile robot in orange color moves from its home location to its designated retrieval rack location (step I in Fig. \ref{problem}). Then it proceeds to lift and transport the corresponding racks to its assigned picking station (step II). Subsequently, after the staff members retrieve the required cargoes at the picking station, the mobile robot relocates the completed rack to the allocated unoccupied storage location (step III). If there still exist retrieval racks remaining, the mobile robot will move towards a newly assigned rack (step IV) and repeats steps II to IV cyclically. On the contrary, if all retrieved racks have been picked up already, then the mobile robot will return home for recharging (step V), and the entire operational process in RMFS comes to an end. 

The challenges in developing an ideal task planner for hyper scale MRTP in RMFS are evident in the following aspects: 

(1) Dimension disaster: the policies for TA and TS in MRTP are coupled and the potential computing time grows exponentially with the scale of elements in the RMFS involving robots, retrieval racks, picking stations and unoccupied storage locations,  which makes traditional task planners difficult to generate a satisfactory solution within effective time.

(2) Dynamic properties: the set of tasks assigned by the top-level order are temporal varying and robots will alter the features of system elements while executing tasks. Besides, the valid elements that can be selected by a robot are also dynamic because robots must obey specific task execution sequences. These dynamic properties augment planning complexity and impose higher requirements on  dynamic response.

These challenges impede our goal of achieving higher quality and faster planning speed for various MRTP in RMFS. 
Some significant related studies have made important contributions to reduce these challenges, the most popular studies mainly involve heuristic methods \cite{bibtex1, camisa2022multi, bibtex3, shi2023bi, bibtex6} and deep reinforcement learning (DRL) methods \cite{ha2021warehouse,bibtex5,ma2024efficient}.
Between them, heuristic methods rely on manual design and their performance are usually limited by the experience of designers.
In contrast, DRL methods can continuously improve their performance through self-exploration, and  solve efficiently considering the strong nonlinear fitting capability of deep networks \cite{mazyavkina2021reinforcement}.
Hierarchical reinforcement learning (HRL) \cite{pateria2021hierarchical}, as a branch of DRL, is considered to have a natural advantage in reducing dimensional disasters for complex tasks while inheriting good dynamic adaptability from DRL.
AlphaStar \cite{vinyals2019grandmaster} outperformed grandmaster human players in the extremely complex game of StarCraft via HRL.
LEE et al. provided an efficient navigation and locomotion method \cite{lee2024learning} for wheeled-legged robot in complex environments through HRL.
These significant works indicate that HRL has the potential to address the challenging MRTP concerned in this article.

However, existing related works still face challenges of scaling up and generalization when dealing with hyper scale MRTP in RMFS.
On the one hand, 
the MRTP in RMFS here is more dependent on global information and optimality, so 
we choose centralized framework instead of decentralized framework \cite{ma2024efficient, camisa2022multi}, and scalability becomes a key challenge beacuse centralized training directly on hyper scale MRTP instances in RMFS with hundreds of robots is still impractical, although HRL can reduce dimension disaster to some extent by decomposing task space.
On the other hand, it is impossible for policies to learn all possible scales and map layouts during training stage, thus excellent generalization performance is indispensable.
In order to overcome all the challenges mentioned, we construct an efficient multi-stage HRL-based multi-robot task planner, and the main contributions are as follows:

(1)	We model the problem of  MRTP in RMFS as a Markov decision process (MDP) with options on asynchronous multi-robot temporal graph with cycle constraints (C2AMRTG), where C2AMRTG is a general extended concept based on temporal graph to abstract the special law of the system.

(2) We construct a centralized hierarchical temporal multi-robot task planning framework following the MDP on C2AMRTG to provide an efficient information updating and interaction mechanism for training and execution. 
The centralized design can ensure optimality, and the hierarchical architecture can reduce the dimension in action space.

(3) We design a hierarchical temporal graph neural network (HTAN) with special temporal embedding layers based on C2AMRTG to enhance spatio-temporal feature extracting performance. It allows basic scaling up ability and can obey varying cycle arc constraints in action space.
 
(4) We propose a HRL algorithm with counterfactual rollout baseline (HCR-REINFORCE) to reduce unfair credit assignment between different policy layers under the hierarchical framework, and use a joint optimization learning technique by combining BC loss with DRL loss to accelerate the convergence speed of learning in the early training stage.

(5) We design a multi-stage curriculum learning method HCR2C by gradually expanding the random boundary of training instances for HCR-REINFORCE to further improve the scaling up and generalization performance while avoiding catastrophic forgetting, and the trained policy can adapt well on various random hyper scale instances and unlearned maps.
After curriculum learning, the policy in our planner can scale up and outperform other heuristic and DRL methods on hyper scale random simulated instances and unlearned maps with up to 200 robots, 1000 retrieval racks and 2000 unoccupied storage racks. Also, our planner successfully outperforms the native company planner on real-world RMFS instances in both planning quality and speed indicators.

Organizational structure: The organizational structure of the rest of this article is as follows. Section \ref{sec:relatework} reviews the related works of MRTP. Section \ref{sec:graph} gives the general concept of C2AMRTG. 
 Section \ref{sec:model} establishes the model of MDP with options on C2AMRTG. Section \ref{sec:framework} introduces the centralized hierarchical temporal multiple robot task planning framework. Section \ref{sec:net} introduces the network HTAN. In Section \ref{sec:train}, the training methods are illustrated in detail. In Section \ref{sec:result}, the effectiveness of our methods are evaluated through comparison experiments. Finally, Section \ref{sec:conclusion} sums up this paper.

\section{Related Works}
\label{sec:relatework}
\subsection{Multi-Robot Task Planning}
Many state-of-the-art works have been done on the sub-problems of MRTP.
In \cite{2021scheduling}, the TS problem for multiple cooperative robot manipulators is studied with a priority constraint optimization model. In \cite{samiei2023distributed}, TA is efficiently addressed in a decentralized fashion across multiple agents with the nearest neighbor search method. In \cite{motes2020multi}, both TA and TD are considered to address the sequential sub-task dependencies in decomposable tasks. In \cite{fu2022robust}, all of the three sub-problems are simultaneously optimized with flexible task decomposing ways.
%The aforementioned approaches addressed the problem within a static planning framework that relies on initial information to plan all future tasks, resulting in a fixed solution lacking adaptability. 
The aforementioned studies either ignore system dynamics, or fail to scale up.
To further improve adaptability, 
\cite{luo2022temporal} focused on exploring a dynamic and scalable framework for TA with distributed structure to adjust planning strategies according to the latest real-time information.  However, there is still a lack of comprehensive research integrally optimizes the complete MRTP problem involving TA, TS and TD in a dynamic and scalable framework, which is a common challenge in many real-world application scenarios.

\subsection{Multi-Robot Task Planning in RMFS}
Due to the complexity of MRTP in RMFS \cite{enright2011optimization}, most of the research has employed various simplifications to reduce solution difficulty. Some researches modeled the simplied MRTP in RMFS as classic scheduling and routing problems or their variants. Ham et al. \cite{schedule} transformed the problem into a standard job-shop scheduling problem. Gharehgozli et al. \cite{bibtex6} modeled the problem as an asymmetric traveling salesman problem with priority constraints \cite{MATSPsolve} considering the intricate rack restorage scenario.
\cite{camisa2022multi} modeled the problem as a pickup-and-delivery vehicle routing problem.
% and proposed a new constraint programming method to solve it.
Some research transformed the complete MRTP to partial sub-problem in MRTP based on specific assumptions.
 Shi et al. \cite{dynamic4} considered the dynamic property but focused solely on TA in RMFS and utilized rules to streamline the planning process.  Others \cite{ha2021warehouse, cheng2024deep} assumed that once rack is brought to a picking station, the robot must deliver it to their original locations without considering the restorage of racks to avoid dealing with TD and dynamic issues. Lian et al. \cite{spatio} addressed the dynamic TS in RMFS with a hierarchical planning framework, but considered only two mobile robots. 
 These existing methods rely on many assumptions and simplifications, which will inevitably lead to the decreased task completion efficiency in RMFS.
Different from the above works,
Zhuang et al. \cite{bibtex3} considered the requirement of rack restorage and delved into both TA and TS in RMFS, but the scale considered only involves 10 robots at most.

\subsection{Solving Multi-Robot Task Planning in RMFS}
The solving time grows exponentially with the scale of problem, considering even the simplified problem of MRTP in RMFS has been proved to be NP-hard or NP-complete \cite{allocation}.
When scale is not very large,
the simplified problem of MRTP in RMFS can be solved precisely with existing mathematical tools \cite{schedule}, such as Gurobi, CPLEX, and SCIP \cite{kleinert2021survey}. 
But for large and hyper scale MRTP in RMFS, it is not realistic to use precise methods to solve on time because they fail to generate a satisfactory solution in a reasonable amount of time.
So many heuristic methods with approximate solution are proposed to accelerate the process of solving. In \cite{camisa2022multi}, heuristic rules together with linear programming method is designed to solve the generalized pickup-and-delivery vehicle routing problem.
Gharehgozli et al. \cite{bibtex6} developed an adaptive large neighborhood search heuristic method to solve efficiently. Ha et al. \cite{ha2021warehouse} used a genetic algorithm with collision evaluation for TS. In \cite{bibtex3}, a heuristic decomposition method within a rolling horizon framework is proposed for  MRTP in RMFS. 
Although these heuristic-based methods can greatly shorten solution time compared to precise methods, they require labor cost to design operators for specific scenarios and still takes a lot of time to achieve a  solution with satisfactory quality.

In recent years, there have been many successful applications of deep reinforcement learning that have demonstrated superior performance to human capabilities, extending from virtual gaming environments to real-world physical domains \cite{vinyals2019grandmaster, kaufmann2023champion}. 
Unlike heuristic-based methods, they do not require continuous search or iterative processes during the reasoning stage, and are capable of providing direct solutions with rapid solving speeds \cite{mazyavkina2021reinforcement}. Furthermore, DRL does not depend on labels and has the potential to yield superior solutions beyond expectations.
Many researchers \cite{koolattention, zhang2020learning, a2c} have observed these advantages of DRL.
\cite{koolattention} provided the method of REINFORCE \cite{REINFORCE} with rollout baseline (R-REINFORCE) to improve solving efficiency for single-agent classical planning problems.
Radosavovic et al. \cite{JPPO} proposed a joint optimization reinforcement learning method JPPO based on PPO \cite{PPO} and supervision learning to improve learning efficiency for complex and highly dynamic location tasks. 
In work \cite{cheng2024deep} and \cite {spatio}, DRL with heuristic rules is used for TA to reduce system costs.
Ho et al. \cite{ho2022federated} adopted a distributed DRL method to realize scaling up.
However, neither of these DRL-based works can solve the generalization challenge on unlearned maps well and successfully verify their methods on hyper scale instances with more than 100 robots and 1000 retrieval racks in RMFS.
In conclusion, there is still a lack of sufficient solving methods for MRTP in RMFS that simultaneously consider scalability, generalization, hyper scale and dynamic challenges. Further works are required to improve planning efficiency for intricate real-world hyper scale MRTP in RMFS.

\section{Asynchronous Multi-Robot Temporal Graph with Cycle Constraints}
\label{sec:graph}
In this section, we introduce the general concept of asynchronous multiple robot temporal graph with cycle constraints (C2AMRTG) to facilitate the description of model, framework and methods in the subsequent sections.

A C2AMRTG consists of a global graph with temporal node features and several robot temporal digraphs. This kind of graph mainly has three kinds of properties: (1) the node information of robot temporal digraphs are updated from the latest global graph asynchronously; (2) the features in global graph and the robot temporal digraphs are time-varying; (3) the set of arcs in robot temporal digraphs have special cycle constraints. The definition for C2AMRTG is given  as follows.

Define the global graph with temporal node features as $\mathcal{G}_g=(\mathcal{V}_g, \mathcal{A}_g)$, where $\mathcal{V}_g=\{v^g_1, v^g_2, \dots, v^g_i, \dots, v^g_N\}$ is the set of nodes, $\mathcal{A}_g$ is the set of global arcs, and the global graph is fully connected.
%, so we are more concerned with the features of nodes in graph $\mathcal{G}_g$. 
Denote the temporal feature vector at the timestamp of $t^g_j$ for a node $v^g_i\in \mathcal{V}_g$ as $f^g_{i,j}=\left(p_x(v^g_{i},t^g_j), p_y(v^g_{i},t^g_j), \mathbb{I}(v^g_{i},t^g_j), \phi(v^g_{i},t^g_j) \right)$, where  $p_x\left(v^g_{i},t^g_j\right)$ and $p_y\left(v^g_{i},t^g_j \right)$ are respectively the position of node $v^g_i$ at the timestamp of $t^g_j$  on the x-axis and y-axis. $\mathbb{I}(v^g_{i},t^g_j)$ is an instructor function with $\mathbb{I}(v^g_{i},t^g_j)=1$ if the node $v^g_{i}$ exists at the timestamp of $t^g_j$, and $\mathbb{I}(v^g_{i},t^g_j)=0$ otherwise. 
Assuming that the set of nodes in global graph $\mathcal{G}_g$ can be further split into $N_{g}+1$ number of subsets $\mathcal{V}_g=\cup_{s=0}^{N_{g}}\mathcal{V}^{g}_s$ where $s$ is the type ID of node.  Then $\phi(v^g_{i},t^g_j) \in \{0,1,\dots,N_g\}$ is the type ID of the node $v^g_{i}$ at the timestamp of $t^g_j$, which means $v^{g}_{i}\in \mathcal{V}^{g}_{\phi(v^g_{i},t^g_j)}$.
%If the current running time $t=t^g_j$, it means the property of the global graph has just changed, and we denote it as a 
Let $\varepsilon_{j}^g$ be a global updating event at the timestamp of $t=t^g_j$. When a global updating event happens, $\mathcal{G}_g$ will be updated. Assuming that event $\varepsilon_{j}^g$ can be completed instantaneously with no time consumption. Events that happened at different time constitute a sequence in accordance with time:  $\varepsilon_{1}^g, \varepsilon_{2}^g, \dots, \varepsilon_{j}^g, \dots, \varepsilon_{T_g}^g$, and all the event elements form a set of global updating event $\mathcal{E}_g$.

Additionally, each mobile robot owns a robot temporal digraph. Let $\mathbb{G}_{r_l}=(\mathbb{V}_{r_l}, \mathbb{A}_{r_l})$ be the robot temporal digraph for the $l$th robot, where $\mathbb{V}_{r_l}=\{\left(v, t_s^v, t_e^v\right)\}$ and $\mathbb{A}_{r_l}=\{((u, v), t_s^a, t_e^a)\}$ are specifically the set of temporal nodes and set of temporal arcs for $\mathbb{G}_{r_l}$. The node $v$ exists from the starting timestamp of $t_s^v$ to the ending timestamp of $t_e^v$ with $t_s^v<t_e^v$ and $t_s^v, t_e^v\in \mathbb{R}_{\ge 0}$. The  directed arc $(u,v)$ exists from the starting timestamp of $t_s^a$ to the ending timestamp of $t_e^a$ with $u, v \in \mathbb{V}_{r_l}$, $t_s^a<t_e^a$ and $t_s^a, t_e^a\in \mathbb{R}_{\ge 0}$.
Let $\mathcal{E}_{r_l}$ be the set of robot graph updating events that will cause $\mathbb{G}_{r_l}$ to change, and the event elements form a sequence $\varepsilon_{1}^{r_l}, \varepsilon_{2}^{r_l}, \dots, \varepsilon_{k}^{r_l}, \dots, \varepsilon_{T_{r_l}}^{r_l}$, where $\varepsilon_{k}^{r_l}$ is the $k$th event that  triggers the graph $\mathbb{G}_{r_l}$ to change. 
%Also, we assume that event $\varepsilon_{k}^{r_l}$ can be completed instantaneously with no time consumption.
Once a robot graph updating event $\varepsilon_{k}^{r_l}$ happens, the $l$th robot will update its robot temporal digraph by pulling the graph node information from the global graph, and we record the updating timestamp as $t^{r_l}_k$.
If we take a shot of the robot temporal digraph $\mathbb{G}_{r_l}$ at each timestamp $t^{r_l}_0, t^{r_l}_1,\dots,t^{r_l}_k, \dots, t^{r_l}_{T_{r_l}}$, we can obtain a set of robot temporal digraph in the form of discrete time $\mathcal{G}_{r_l}=\{\mathcal{G}^{r_l}_0, \mathcal{G}^{r_l}_1, \dots, \mathcal{G}^{r_l}_k, \dots, \mathcal{G}^{r_l}_{T_{r_l}}\}$, where $\mathcal{G}^{r_l}_k=\left(\mathcal{V}^{r_l}_k, \mathcal{A}^{r_l}_k\right)$ is a robot digraph shot. $\mathcal{V}^{r_l}_k=\{v^{r_l}_1, v^{r_l}_2, \dots, v^{r_l}_i, v^{r_l}_N\}$ and $\mathcal{A}^{r_l}_k=\{(u,v)|u,v\in \mathcal{V}^{r_l}_k\}$ are respectively the set of nodes and the set of arcs in discrete time for the $l$th robot. At the timestamp of $t^{r_l}_k$, both the set of nodes $\mathcal{V}^{r_l}_k$ and the node feature vector $f^{r_l}_{i,k}$ for the $l$th robot are directly updated from the current global graph $\mathcal{G}_g$, which means $v^{r_l}_i=v^g_i$, $f^{r_l}_{i,k}=\left( p_x(v^{r_l}_{i},t^{r_l}_k), p_y(v^{r_l}_{i},t^{r_l}_k), \mathbb{I}(v^{r_l}_{i},t^{r_l}_k), \phi(v^{r_l}_{i},t^{r_l}_k) \right)$, and $f^{r_l}_{i,k}= f^{g}_{i,j}$. To be more specific, $p_x(v^{r_l}_{i},t^{r_l}_k)=p_x(v^{g}_{i},t^{g}_j)$, $p_y(v^{r_l}_{i},t^{r_l}_k)=p_y(v^{g}_{i},t^{g}_j)$, $\mathbb{I}(v^{r_l}_{i},t^{r_l}_k)=\mathbb{I}(v^{g}_{i},t^{g}_j)$, and $\phi(v^{r_l}_{i},t^{r_l}_k)=\phi(v^{g}_{i},t^{g}_j)$, where $t^{g}_j$ is the latest global graph updating time before $t_{k}^{r_l}$.

Let $\mathcal{H}^{r_l}_k=(h^{r_l}_{0}, h^{r_l}_{1}, \dots, h^{r_l}_{i}, \dots, h^{r_l}_{k})$ be the historical node ID sequence for the $l$th robot before $t_k^{r_l}$, where $h^{r_l}_{i}\in \mathbb{N}$ is the ID of the last node that was assigned for the $l$th robot before the timestamp of $t_i$. 
It is obviously that the set of nodes in $\mathbb{G}_{r_l}$ can also be further split into $N_{g}+1$ number of subsets  $\mathbb{V}_{r_l}=\cup_{j=0}^{N_{g}}\mathbb{V}^{r_l}_j$. 
Let $\hat{\sigma}_{r_l}:= \left( \begin{matrix}
	1& 2 & \dots & N_{g}\\
	2& 3 & \dots & 1
\end{matrix} \right)$ be the cycle \cite{kostrikin1982introduction} of the $l$th robot to represent the cyclic visiting sequence that the $l$th robot has to follow. It is essentially a mapping from set $\{1,2,\dots,N_{g}\}$ consisting of $N_{r_l}$ number of subset ID elements to itself $\hat{\sigma}_{r_l}:j\mapsto \hat{\sigma}_{r_l}(j)$, and two cycles obtained by shifting are same.

	We use an adjacency matrix $\mathcal{M}^{r_l}_{k}$ whose element is $m^{r_l,k}_{i,j} \in \{0,1\}^{|\mathcal{A}^{r_l}_k|}$  to represent the connectivity relationship starting from 
	node $v^{r_l}_{i} \in \mathcal{V}^{r_l}_k$ to  node $v^{r_l}_{j} \in \mathcal{V}^{r_l}_k$ for $\mathcal{A}^{r_l}_k$. If  $m^{r_l,k}_{i,j}=1$, the $l$th robot can travel from the node $v^{r_l}_{i}$ to the node $v^{r_l}_{j}$. Otherwise, the $l$th robot cannot travel from the node $v^{r_l}_{i}$ to the node $v^{r_l}_{j}$.
	If a historical node ID sequence $H^{r_l}_k$ is given and $k\geq N_{r_l}$, 
	then for any arc $\left(v^{r_l}_{i}, v^{r_l}_{j}\right)\in \mathcal{A}^{r_l}_k$ at the timestamp of $t^{r_l}_k$, 
	it must follow the robot arc constraints as follows:
	
	(1) $\mathbb{I}(v^{r_l}_{i},t^{r_l}_k)=1$, and $\mathbb{I}(v^{r_l}_{j},t^{r_l}_k)=1$ (node existence constraint);
	
	(2) $h^{r_l}_{k}=i$, and $h^{r_l}_{k+1}=j$ (spatiotemporal consistency constraint);

	(3) $\phi(v^{r_l}_{h^{r_l}_{k+1}},t^{r_l}_{k+1}) =\hat{\sigma}_{r_l}(\phi(v^{r_l}_{h^{r_l}_{k}},t^{r_l}_{k})))$ if $1\leq k< T_{r_l}$ (cycle constraint);

	(4) $\phi(v^{r_l}_{h^{r_l}_{0}},t^{r_l}_{0}) = 0$, $\phi(v^{r_l}_{h^{r_l}_{1}},t^{r_l}_{1}) = 1$ and $\phi(h^{r_l}_{T_{r_l}}) = 0$ (terminal constraint).
	
	If the arc $\left(v^{r_l}_{i}, v^{r_l}_{j}\right)$ satisfies all of the constraints above, then $m^{r_l,k}_{i,j}=1$, and $m^{r_l,k}_{i,j}=0$ otherwise. 
	A practical physical explanation for these constraints is the robot must start from the initial node and circularly visit the nodes in relative type node sets in a fixed cyclic sequence. Once the robot completes the required task, it must return to its initial node.

\section{Model Formulation} 
\label{sec:model}
In this section, we assign specific physical meanings to variables in C2AMRTG and introduce the MDP model with options on C2AMRTG to describe MRTP in RMFS. 

To facilitate the analysis and solution of the problem, we specify the following assumptions within a reasonable range:

(1) Each robot can only carry one rack at a time.

(2) Each unoccupied storage location can only be used to store just one rack.

(3) The traveling time consumed only corresponds to the Manhattan distance traveled and average robot moving speed. 

(4) The time of picking, queuing, lifting up and putting down a rack for robots can be ignored.

The C2AMRTG has specific parameters and physical meanings for representing problem of MRTP in RMFS with these assumptions. In the global graph $\mathcal{G}_g$, we can split the set of nodes $\mathbb{V}_{g}$ into four subsets with $N_g=3$, which can be written as $\mathbb{V}_{g}=\cup_{j=0}^{3}\mathbb{V}^{g}_j$, where  
$\mathbb{V}_0^{g}=\{v^{g}_1, v^{g}_2, \dots, v^{g}_{N_a}\}$ is the set of robot home nodes, and $v^{g}_i$ is the robot home node for the $i$th robots with $i\in\{1,2,\dots,N_a\}$.  $\mathbb{V}_1^{g}=\{v^{g}_{N_a+1}, v^{g}_{{N_a}+2}, \dots, v^{g}_{N_a+N_r}\}$ is  the set of retrieval rack nodes, $\mathbb{V}_2^{g}=\{v^{g}_{N_a+N_r+1}, v^{g}_{N_c+N_r+2}, \dots,  v^{g}_{N_a+N_r+N_p}\}$ is the set of picking station nodes, and  $\mathbb{V}_3^{g}=\{v^{g}_{N_a+N_r+N_p+1}, v^{g}_{N_a+N_r+N_p+2}, \dots,  v^{g}_{N_a+N_r+N_p+N_s}\}$ is the set of unoccupied storage location nodes. 
The total number of nodes in the global graph is $N_w=N_a+N_r+N_p+N_s$.  Each node in the set of $\mathbb{V}_{g}$ just corresponds to a possible delivery destination location.  
The type ID property of the node in the global graph at the initial moment is defined as:
\begin{equation}
	\phi(v_i^g,t_0^g) = 
	\begin{cases}
		0,  &\text{if $1\leq i\leq N_a$;} \\
		1,  &\text{if $N_a< i\leq N_a+N_r$;} \\
		2,  &\text{if $N_a+N_r< i\leq N_a+N_r+N_p$;} \\
		3,  &\text{if $N_a+N_r+N_p\leq i\leq N_w$.} 
	\end{cases}
\end{equation}

All nodes exist at the initial moment with $\mathbb{I}(v^g_{i},t^g_0)=1$.
The whole task in RMFS can be broken down into a series of delivery sub-tasks and each delivery sub-task just corresponds to a node in  $\mathbb{V}_{g}$ according to its terminal location.
When a mobile robot in the system is assigned with a node or the assigned robot just completed a delivery sub-task, the global updating event  $\varepsilon_{j}^g \in \mathcal{E}_g$ happens, and the global graph will be updated.
The  global updating event includes three kinds of cases: (1) If a  robot is assigned with a node involving delivering a picked rack to the allocated unoccupied storage location, the related unoccupied storage location node will update its existence instructor function with $\mathbb{I}(v^g_{i},t^g_j)=0$. 
(2) If a retrieval rack is assigned to a robot, the related retrieval rack node will update its existence instructor function with $\mathbb{I}(v^g_{i},t^g_j)=0$ to prevent repeated rack assignment. 
(3) If the robot arrives at the location of the assigned retrieval rack, the related retrieval rack node will update its type ID with $\phi(v_i^g,t_j^g) = 3$, and the related retrieval rack node will update its existence instructor function with $\mathbb{I}(v^g_{i},t^g_j)=1$.

%There are $N_a$ number of robot temporal digraph $\mathbb{G}_{r_l}=(\mathbb{V}_{r_l}, \mathbb{A}_{r_l})$.
Let $\mathcal{E}_{R}=\{\varepsilon^R_{0}, \varepsilon^R_{1},\dots, \varepsilon^R_{m},\dots, \varepsilon^R_{T_R^{r_l}}\}$ be the set of robot selecting decision events.
A robot selecting decision event $\varepsilon^R_{m} \in \mathcal{E}_{R}$ for MRTP will be triggered when there exists a robot that has no uncompleted delivery sub-task any more. Assuming that the $l$th robot is selected to assign a node, then the robot graph updating event $\varepsilon_{k}^{r_l} \in \mathcal{E}_{r_l}$ will be triggered and the node features in $\mathbb{G}_{r_l}$ are updated directly from the $\mathcal{G}_g$ at $t_j^g$. Let $\mathcal{H}_{j}^a=(h^a_0,  h^a_1, \dots, h^a_{j})$ be the history sequence of selected robot IDs until $t_j^g$.
The related discrete robot digraph shot is $\mathcal{G}^{r_l}_k=(\mathcal{V}^{r_l}_k, \mathcal{A}^{r_l}_k)$. 
The cycle constraint in $\mathcal{G}^{r_l}_k$ represents the visiting sequence as follows:
(a) firstly, the robot should visit a retrieval rack node; 
(b) then it should visit a picking station node;
(c) finally, it should visit a unoccupied storage location node.
So $N_g=3$, $\hat{\sigma}_{r_l}:= \left( \begin{matrix}
	1& 2 &  3\\
	2& 3 & 1
\end{matrix} \right)$, and any arc in the  robot digraph shot  should follow all the four kinds of robot arc constraints in C2AMRTG.
Except for these default constraints, the problem should also obey a special constraint that the retrieval rack should be delivered to its related 
picking station, which can be written as $ h_{k+1}^{r_l}=f(h_k^{r_l})$ if $\phi(v_{h_k^{r_l}}^{r_l}, t_k^{r_l})=2$, and $f(\cdot)$ is a mapping from the set of retrieval rack node ID to the set of picking station node ID.

Based on the specific C2AMRTG defined above, the optimization objective of MRTP problem is to minimize the makespan until all the delivery sub-tasks are completed.
Here, makespan is the total time consuming until the task termination condition is established, and the task termination condition for the RMFS system is defined as: all the retrieval racks are delivered by robots to the required picking stations and put to an unoccupied storage location, and all the robots go back to their home locations for recharging.

\subsection{MDP with Options on C2AMRTG}
According to these definitions, the MRTP problem in RMFS can be modeled as MDP with options on C2AMRTG based on the work of \cite{sutton1999between} by augmenting MDP with a predetermined set of options to facilitate the utilization of reinforcement learning methodology. 
The model of MDP with options can be described by a tuple $(S,O,P,R,F)$, where $S$ is the set of all possible states, which includes $\mathcal{G}_g$, $\mathbb{G}_{r_l}$, $\mathcal{H}_{j}^a$ and $\mathcal{H}^{r_l}_k$. $O$ is the set of options. $ P:S\times O\times S \mapsto [0,1]$ is the set of state and option transition probability function and it follows the updating rules of C2AMRTG for MRTP in RMFS defined above. $R$ is an instant reward function. 
$ F:S\times O\times S\mapsto (\mathbb{R}_{\ge 0}\mapsto [0,1])$ is a continuous probability distribution function giving probability of transition times for each state-option pair, it is determined by the policy and termination condition defined in options.

Option can be viewed as the generalization of primitive action with time extensibility.
In the set of options $O$, 
each option element $o$ can be described by $\left<I, \pi, \beta \right>$, where $I \subset S$ is an initiation state set and $\pi: S_o\times A_o \mapsto [0,1]$ is the policy. $S_o\subset S$ is the state set of $\mathcal{O}$ and $A_o\subset A$ is the action set of $\mathcal{O}$. $\beta:S_o^{+}\mapsto [0,1]$ is the termination condition, indicating the option will terminate at each time step following the probability of $\beta$ after an execution from the initial state based on policy $\pi$. $S_o^{+}$ is the set of states including termination state, and we define the termination state set for $\mathcal{O}$ as $S_o^{\dag}$.

Considering the whole MRTP in RMFS consists of three sub-problems including TA, TS and TD,
we break the set of options $\mathcal{O}$ into two parts: 
$\mathcal{O}=\mathcal{O}_{r}\cup \mathcal{O}_{g}$, where $\mathcal{O}_{r}$ is the set of robot selecting option and it is related to TA.
$\mathcal{O}_{g}$ is the set of graph node selecting option and it is related to TS. 
Furthermore, we can split the node location option set $\mathcal{O}_{g}$ into disjoint subsets $\mathcal{O}_g=\cup_{j=0}^{3} \mathcal{O}^g_{j}$ according to the decomposition mode in $\mathbb{V}_{r_l}$, and this  process is related to TD.
 Among them,  $\mathcal{O}^g_{0}$ is the set of robot home option related to selecting the required robot home node, $\mathcal{O}^g_{1}$ is the set of retrieval rack options related to selecting a rack node, $\mathcal{O}^g_{2}$ is the set of picking station options related to selecting a picking station node, and $\mathcal{O}^g_{3}$ is the set of unoccupied storage options related to selecting an unoccupied storage location node. 
Considering the two kinds of options are generated in succession, we define a joint option $o_m=(o^r_m,o^g_m)$ to represent the sequential execution of robot option and graph node option at the $m$th joint option decision step, where $o^r_m\in \mathcal{O}_{r}$ is the robot option at the $m$th robot option decision step and $o^g_m\in \mathcal{O}_{g}$ is the graph node option at the $m$th graph node option decision step.
The application method of joint option $o_m$ is similar to the action in traditional Markov decision process framework, but we should notice that the duration of action is fixed.
The detailed definitions for each option set are explained as follows.

\subsection{Robot Selecting Option}
In the set of robot selecting options $\mathcal{O}_{r}=\left<I_r, \pi_r, \beta_r \right>$, 
the initial state set $I_{r} \subset S_r$ encompasses all states $s^r_m$ when an event $\varepsilon_{m}^R \in \mathcal{E}_{R}$ happens. Specifically, $s^r_m$ consists of the node features $f^g_{i,j}$ of global graph snapshots $\mathcal{G}_g$ when an event $\varepsilon^g_j \in \mathcal{E}_g$ has just happened before $\varepsilon_{m}^R$ happened. 
Policy $\pi_{r}: S_r\times A_r \mapsto [0,1]$ outputs the probability distribution of robot selecting action through state $S_r$, where $S_r$ is composed of $\mathcal{G}_g$ and $\mathcal{H}_{j-1}^a$. The robot selecting action set $A_r$ includes the IDs of all the robots that have not completed their delivery sub-tasks and returned home.
$\beta_{r}$ is the robot selecting option termination condition. Let $S_{r}^{\dag}$ be the termination state set of $\mathcal{O}_{r}$, then $\beta_{r}(S_{r}^{\dag})=1$ means that option ends up immediately when  the system completes an action of selecting a robot.

\subsection{Graph Node Selecting Option}
For the set of graph node selecting options $\mathcal{O}_{g}=\left<I_g, \pi_g, \beta_g \right>$, 
the initial state set $I_{g}\subset S_g$ encompasses all states $s^g_m$ when $\varepsilon_{k}^{r_l} \in \mathcal{E}_{r_l}$ is triggered, and this means it is the turn for system to select a graph node which is the goal location for the latest robot $l=h^a_{j-1}$ assigned before $t=t^{r_l}_k$. 
Policy $\pi_{g}: S_g\times A_g \mapsto [0,1]$ outputs the probability distribution of the graph node  selecting action through state $S_g$, where $S_{g}$ is the set of graph node states. Specifically, it consists of 
 $\mathcal{G}_{r_{l}}$, 
 $\mathcal{H}_{j}^a$, and $\mathcal{H}^{r_l}_k$ 
  until  $t=t^{r_{l}}_k$. 
Because the set $I_{g}$, $A_g$ and the node selecting option termination condition $\beta_{g}$ differs for different subsets $\mathcal{O}^{g}_0, \mathcal{O}^{g}_1, \mathcal{O}^{g}_2, \mathcal{O}^{g}_3$ in $\mathcal{O}_{g}$, we explain each type of graph node selecting option in further detail:

\subsubsection{{Robot Home Option $\mathcal{O}^g_{0}$}}
$I^g_{0}$ is the initial set of states in $\mathcal{O}^g_{0}$, encompassing all states  when an event in $\mathcal{E}^h_l$ happens. If the retrieval racks have all been carried to the related picking stations and the robot $r_l$ has delivered its last rack to an unoccupied storage location, the selected $l$th robot should return to its home location (step V in Fig. \ref{problem}), and we denote the set of these events as $\mathcal{E}^h_l$. Home action set $A^g_{0}$ consists of only one action: selecting its home node $v_l^{r_l}$. $\beta^g_{0}$ is the termination condition. Let $S_{0}^{\dag}$ be the termination state set of $\mathcal{O}^g_{0}$, then  $\beta ^g_{0}(S_{0}^{\dag})=1$ means option ends up immediately when robot $r_l$ arrives at location of $v_l^{r_l}$ or an event $\varepsilon_{m}^R \in \mathcal{E}_{R}$ happens. 

\subsubsection{Retrieval Rack Option $\mathcal{O}^g_{1}$}
The initial set of states in $\mathcal{O}^g_{1}$, denoted as $I^g_{1}$, encompasses all states when it is the turn for the $l$th robot to make a decision of selecting a retrieval rack (step I and IV in Fig. \ref{problem}). Retrieval rack action set $A^g_{1}$ consists of actions of selecting a valid retrieval rack node $v_j^{r_l} \in \mathcal{V}_{r_l}$ with the feature $\phi(v_j^{r_l},t_k^{r_l})=1$ and $m_{{h_k^{r_l}},j}^{r_l,k}=1$. Let $S_{1}^{\dag}$ be the termination state set of  $\mathcal{O}^g_{1}$, then $\beta^g_{1}(S_{1}^{\dag})=1$ means option ends up immediately when the system completes an action of selecting a retrieval rack node or an event $\varepsilon_{m}^R \in \mathcal{E}_{R}$ happens.

\subsubsection{Picking Station Option $\mathcal{O}^g_{2}$}
The initial set of states in $\mathcal{O}^g_{2}$, denoted as $I^g_{2}$, encompasses all states when it is the turn for the $l$th robot to make a decision of selecting a picking station (step II in Fig. \ref{problem}). Picking station action set $A^g_{2}$ consists of the actions of selecting a valid picking station node $v_z^{r_l}\in \mathcal{V}^2_{r_l}$ with features of $\phi(v_z^{r_l},t_k^{r_l})=2$ according to the latest retrieval rack that has been assigned for the $l$th robot, which means $z=f(h_{k}^{r_{l}})$, and $z$ is the current valid picking station node ID for the $l$th robot.  Let $S_{2}^{\dag}$ be the termination state set of $\mathcal{O}^g_{2}$, then $\beta ^g_{2}(S_{2}^{\dag})=1$ means that option ends up immediately when the assigned robot arrives at the location of selected picking station or an event $\varepsilon_{m}^R \in \mathcal{E}_{R}$ happens.

\subsubsection{Unoccupied Storage Option $\mathcal{O}^g_{3}$}
The initial set of states in $\mathcal{O}^g_{3}$, denoted as $I^g_{3}$, encompasses all states when it is the turn for the $l$th robot to make a decision of selecting an unoccupied storage location (step III in Fig. \ref{problem}). Unoccupied storage action set $A^g_{3}$ consists of actions of selecting an unoccupied storage location node $v_j^{r_l}$ with features of $\phi(v_j^{r_l},t_k^{r_l})=3$ and $m_{{h_k^{r_l}},j}^{r_l,k}=1$ from $\mathcal{V}_{r_l}$.  Let $S_{3}^{\dag}$ be the termination state set of $\mathcal{O}^g_{3}$, then  $\beta^g_{3}(S_{3}^{\dag})=1$ means that option ends up immediately when the $l$th robot arrives at the location of assigned unoccupied storage location or an event $\varepsilon_{m}^R \in \mathcal{E}_{R}$ happens.

\subsection{Reward}

To calculate the instant reward, we first define the robot traveling time $r_{k,l}$ with a recurrence formula in eq. \eqref{con:2}:
\begin{equation}
	r_{k,l} = 
	\begin{cases}
		&\nu_r*\left(\left|p_x\left(v^{r_l}_{h^{r_l}_{1}},t^{r_l}_{1}\right)-p_x\left(v^{r_l}_{h^{r_l}_{0}},t^{r_l}_{0}\right)\right| \right.\\[1.5ex]
		&\hspace{13pt}+\left.\left|p_y\left(v^{r_l}_{h^{r_l}_{1}},t^{r_l}_{1}\right)-p_y\left(v^{r_l}_{h^{r_l}_{0}},t^{r_l}_{0}\right)\right|\right), \text{if $k=0$;} \\[1.5ex] 
		&r_{m-1,l}+ \\[1ex]
		&\nu_r*\left(\left|p_x\left(v^{r_l}_{h^{r_l}_{k+1}},t^{r_l}_{k+1}\right)-p_x\left(v^{r_l}_{\max(0,h^{r_l}_{k})},t^{r_l}_{k}\right)\right|\right. \\[1.5ex]
		&\hspace{13pt}+\left.\left|p_y\left(v^{r_l}_{h^{r_l}_{k+1}},t^{r_l}_{k+1}\right)-p_y\left(v^{r_l}_{\max(0,h^{r_l}_{k})},t^{r_l}_{k}\right)\right|\right), \\[1.5ex]
		&\hspace{108pt}\text{if $k>0$, and $ \varepsilon_k^{r_l} \notin \mathcal{E}^h_l$;} \\[1.5ex]
		&r_{k-1,l} +\\[1ex]
		&\nu_r*\left(\left|p_x\left(v^{r_l}_{h^{r_l}_{k+1}},t^{r_l}_{k+1}\right)-p_x\left(v^{r_l}_{\max(0,h^{r_l}_{k})},t^{r_l}_{k}\right)\right|\right.\\[1.5ex]
		&\hspace{15.5pt}+\left.\left|p_y\left(v^{r_l}_{h^{r_l}_{k+1}},t^{r_l}_{k+1}\right)-p_y\left(v^{r_l}_{\max(0,h^{r_l}_{k})},t^{r_l}_{k}\right)\right|\right.\\[1.5ex]
		&\hspace{15.5pt}+\left.\left|p_x\left(v^{r_l}_{h^{r_l}_{0}},t^{r_l}_{0}\right)-p_x\left(v^{r_l}_{h^{r_l}_{k}},t^{r_l}_{k}\right)\right|\right.\\[1.5ex]
		&\hspace{15.5pt}+\left.\left|p_y\left(v^{r_l}_{h^{r_l}_{0}},t^{r_l}_{0}\right)-p_y\left(v^{r_l}_{h^{r_l}_{k}},t^{r_l}_{k}\right)\right|\right),\\[1.5ex]
		&\hspace{108pt}\text{if $k>0$ and $ \varepsilon_k^{r_l} \in \mathcal{E}^h_l$.}
	\end{cases}\label{con:2}
\end{equation}
where  $\nu_r$ is the average moving velocity of the mobile robot, and
$\mathcal{E}^h_l$ is the set of events when the $l$th robot should go back home.
For each robot, the robot traveling time $r_{k,l}$ is the total consuming time for the $l$th robot to complete its $k$th delivery sub-task assigned after the $m$th joint option has just been decided. 
Then the definition of the instant reward function is given in eq. \eqref{con:3} over the joint option $o_m$.
\begin{equation}
	\begin{aligned}
		R_m = -(\Phi_{m}-\Phi_{m-1})	
	\end{aligned}\label{con:3}
\end{equation}

where $\Phi_{m}$ is the maximum number of robot traveling time that the $m$th joint option has just been completed.
Based on eq. \eqref{con:2},
$\Phi_{m}$ can be further defined as: $\Phi_{m} = \max(r_{k_{1,m},1}, r_{k_{2,m},2}, \dots, r_{k_{l,m},l}, \dots, r_{k_{N_a,m},N_a})$, where $k_{l,m}$ is the number of assigned nodes for the $l$th robot when the $m$th joint option has just been decided. 
It is obvious that makespan just equals to the maximum number of $r_{k,l}$ when tasks are all terminated, so  discount factor $\gamma=1$ and epoch reward $\sum_{m=0}^{\max_l{(T_R^{r_l})}}{\gamma^mR_{m}}$ equals to makespan.

\section{Hierarchical Temporal Multi-Robot Task Planning Framework} 
\label{sec:framework}
\begin{figure*}[!t]
	\centering
	\includegraphics[width=6.6in]{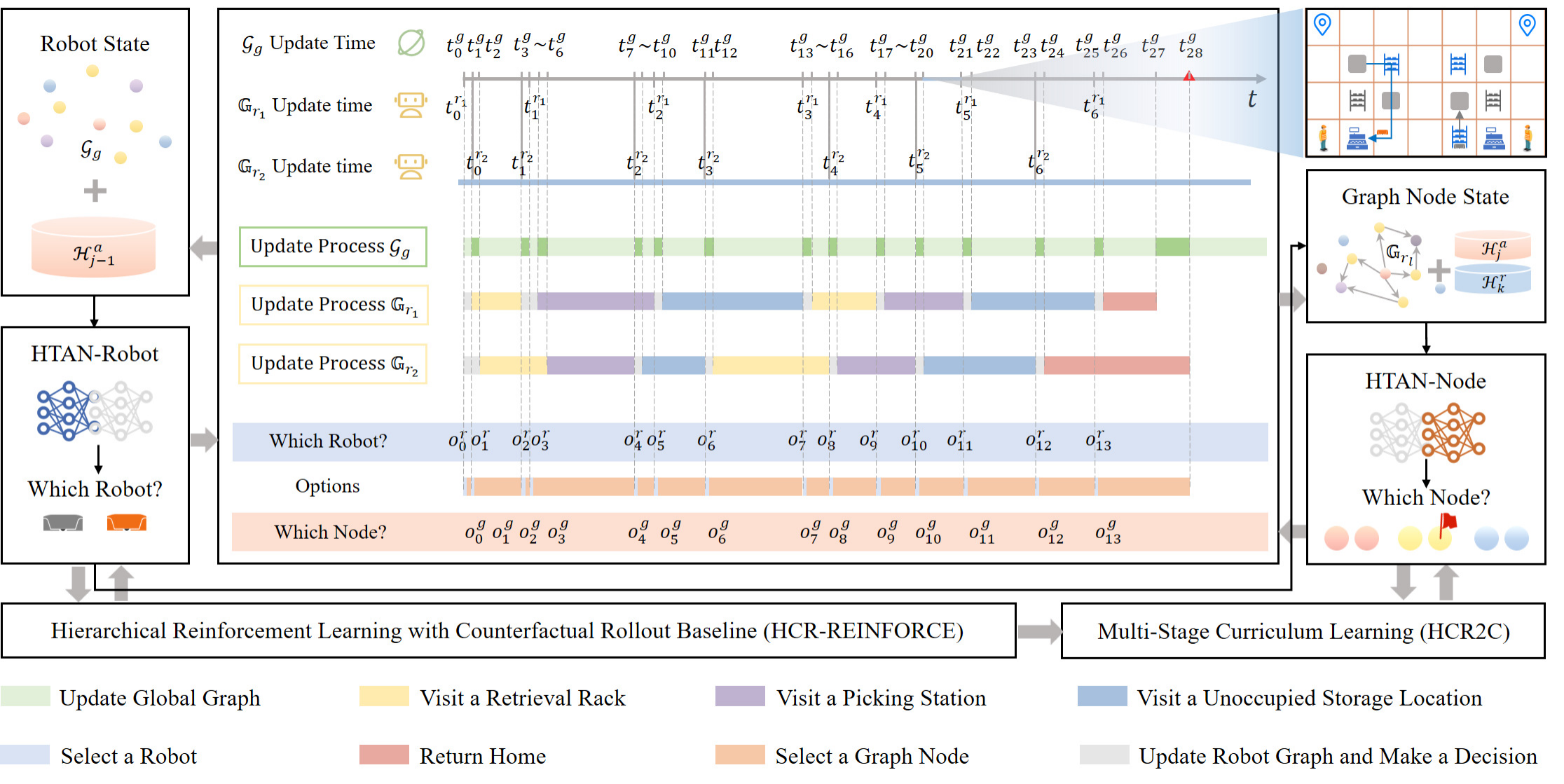}
	\caption{The hierarchical temporal multi-robot task planning framework and temporal logic for a specific multi-robot task planning instance in RMFS with $2$ mobile robots and $4$ retrieval racks.}
	\label{fig_10}
\end{figure*}

In this section, we briefly introduce the whole hierarchical temporal multiple robot task planning framework, 
and show the temporal logic of dynamic planning for MRTP under this planning framework with a simple example.

The hierarchical temporal multiple robot task planning framework is shown in Fig. \ref{fig_10}. It is designed as a centralized  optimization architecture to ensure optimality. Also, it has a hierarchical dynamic planning structure with two layers: (i) the higher layer makes a decision of robot selecting option through the network of  HTAN-robot, (ii) the lower layer then makes a  decision of graph node selecting option for the assigned robot decided in the higher layer through the network of HTAN-node. 
The hierarchical structure can effectively reduce dimension in action space from $n^m$ to $n+m$ assuming there are $n$ robots and $m$ nodes in total.
Based on this hierarchical structure, joint options are generated by the two-layer network HTAN illustrated in Section \ref{sec:net} and they will interact with the system, resulting in graph updating events and decision events. 
%The process will repeat until the task termination condition is established. 

In the process of training, the parameters in HTAN are optimized integrally with a HRL method HCR-REINFORCE to train policies with hierarchical structure efficiently. 
Then, multi-stage curricula are prepared for the HCR-REINFORCE to improve scaling up and generalization performance for unlearned instances by gradually increasing the difficulty and randomness of training instances.
When used for execution, the framework will remove all the training modules, directly using the learned parameters in HTAN to generate options and interact with the  task planning system. 

The temporal logic of decision and graph updating  in our framework follows the asynchronous update characteristics of C2AMRTG. The time intervals between different updating and decision events are unfixed and each robot owns proprietary  observations. 
For easy understanding, we visualize the system temporal logic under our framework in Fig. \ref{fig_10} with an example of MRTP in RMFS with $2$ robots and $4$ retrieval racks.

In Fig. \ref{fig_10}, robot $r_1$ obtains its initial robot digraph shot $\mathcal{G}^{r_1}_0$ at timestamp of $t_0^{r_1}$ with the same node features in the global graph at $t_0^{g}$. Then it is assigned with a retrieval rack at $t_0^{r_1}$, and the existence instructor function for the retrieval rack node in the global graph $\mathcal{G}_{g}$ is updated to nonexistence. At this time, the retrieval rack can not be delivered by other robots. Since robot $r_2$ has not yet been assigned a node, $\varepsilon_{m}^R$ is triggered at $t_1^{g}$ and robot $r_2$ is selected according to the latest global graph. Then robot digraph shot $\mathcal{G}^{r_{2}}_0$ is updated at $t_0^{r_2}$ and $r_2$ is assigned with a retrieval rack. Robot $r_1$ drives from the home location starting at $t_1^g$ and arrives at its assigned retrieval rack location at $t_3^g$, then the type of the selected node in the global graph $\mathcal{G}_{g}$ will be updated to $\phi(v_{h_1^{r_1}}^{g},t_3^{g})=3$. Also, its existence instructor function will be updated to existence. Then the location can be used to store the racks that have been picked. The event $\varepsilon_{m}^R$ is triggered at $t_3^g$. Then robot $r_2$ is selected and assigned with a picking station node according to the latest robot digraph shot $\mathcal{G}^{r_{2}}_1$ at $t^{r_2}_1$. $\mathcal{G}_{g}$ is updated. Soon afterwards, robot $r_1$ is selected at $t_4^g$, and it is assigned with a picking station node at $t_1^{r_1}$ according to $\mathcal{G}^{r_1}_1$. The remain updating and decision process will keep on until the task termination condition defined in \ref{sec:model} is established. 
Through the temporal logic depicted in Fig. \ref{fig_10}, we can find that only decision transitions at the critical timestamps are worthy of attention.  %with unfixed time intervals. 
Obviously, when the system can satisfy immediate dynamic response, this design is more conducive to reducing interaction steps compared to that in synchronous mechanism with fixed time intervals. So this framework is beneficial for training efficiency and system dynamic response.

\section{Hierarchy Temporal Attention Network}
\label{sec:net}
\begin{figure*}[!t]
	\centering
	\includegraphics[scale=0.36]{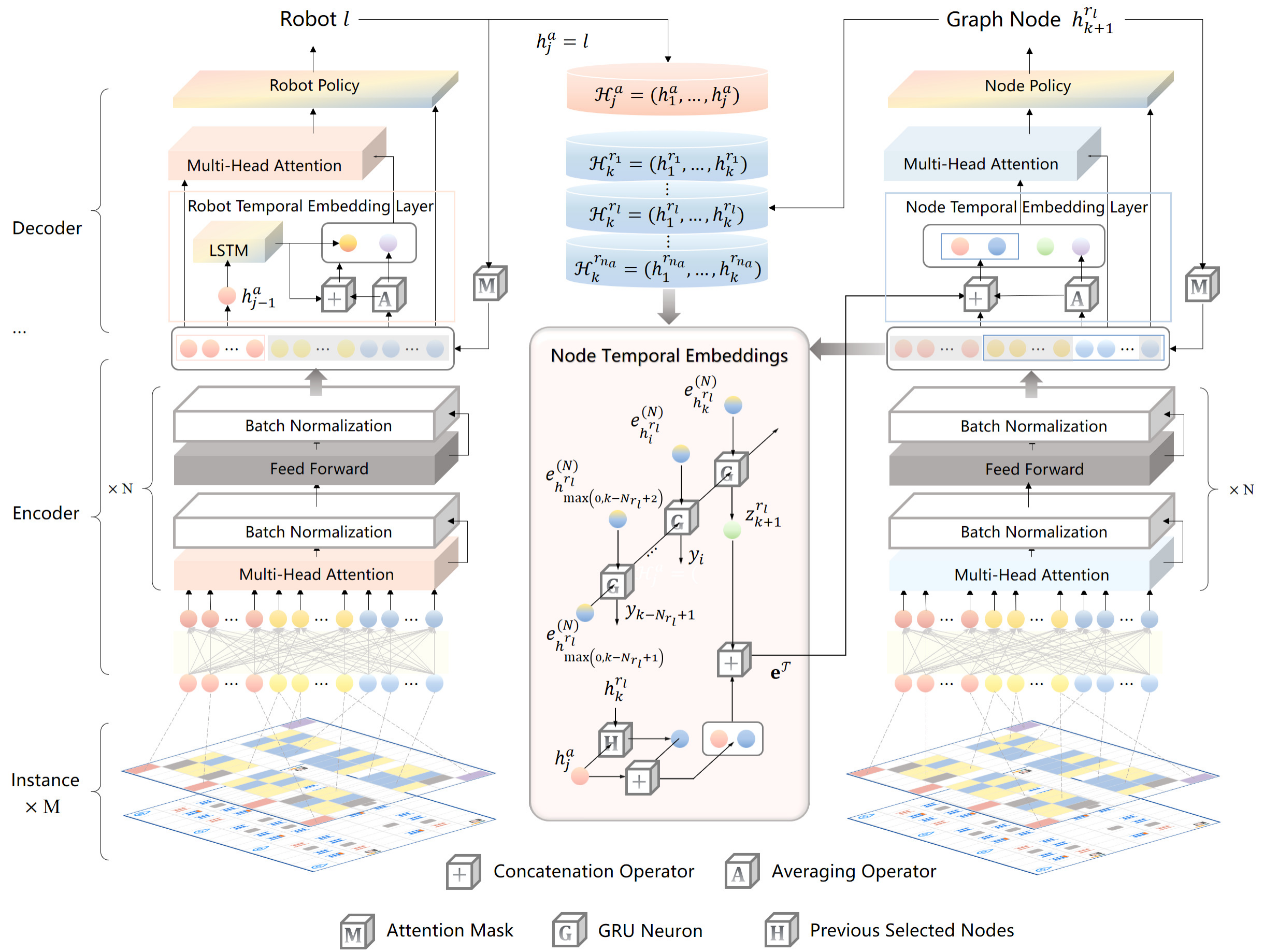}
	\caption{The hierarchy temporal attention network architecture including the robot net (left) and the graph node net (right).}
	\label{network}
\end{figure*}

The hierarchy temporal attention network (HTAN) has two sequentially connected nets,  respectively named robot net (HTAN-robot) and node net (HTAN-node) as shown in the left and right halves of Fig. \ref{network}. 
Both of the robot net and node net have an encoder-decoder structure.
The encoder is built to encode inputs corresponding to the observed global graph and extract  spatial features from nodes. The decoder is designed to further extract temporal features and notable spatial features on the basis of the node embedding generating from encoder. HTAN takes advantage of scalability in attention network and is able to handle variable-length inputs and output probability distribution for varying option sets. 

\subsection{Input}
The input vector ${\bf{x}}=(x_1, x_2,\dots, x_{N_w})$ of robot net and node net is derived from the observed state when relative event is triggered. ${\bf{x}}$ is composed of robot feature vector and global temporal graph node feature vector. 
In detail, robot feature vector contains robot traveling time data, current position vector and home position vector of each robot:
 ${\bf{x}}_l=(r_{k,l},p_x(v^g_{l},t^g_k), p_y(v^g_{l},t^g_k),p_x(v^g_{h_k^{r_l}},t^g_k), p_y(v^g_{h_k^{r_l}},t^g_k))$. Graph node feature vector contains the position vector of each graph node in global graph $\mathcal{G}_g$: 
${\bf{x}}_n=(p_x(v^g_{h_k^{r_l}},t^g_k), p_y(v^g_{h_k^{r_l}},t^g_k),p_x(v^g_{h_k^{r_l}},t^g_k), p_y(v^g_{h_k^{r_l}},t^g_k))$. Specially, if the node corresponds to a retrieval rack $\phi(v^g_{h_k^{r_l}},t_k^g)=1$ and the node is existent $\mathbb{I}(v^g_{h_k^{r_l}},t^g_k)=1$, then ${\bf{x}}_n=( p_x(v^g_{h_k^{r_l}},t^g_k), p_y(v^g_{h_k^{r_l}},t^g_k),p_x(v^g_{f(h_k^{r_l})},t^g_k), p_y(v^g_{f(h_k^{r_l})},t^g_k))$. In this case, picking station nodes are merged with the retrieval rack nodes to reduce input dimension. Also, the type features can be reflected by indexes of nodes
and masks in decoders will deal specifically with the varying type features, so no extra type ID data is provided in the input vector.

\subsection{Encoder}
The encoder structure in the robot net and node net is same as that in transformer encoder \cite{vaswani2017attention}. It is mainly composed of $N$-layer sequentially stacked multi-head attention network and feedforward layer, except for the input layer in encoder which is just a fully-connected linear layer to obtain high-dimension input embedding ${\bf{e}}^{(0)}=(e^{(0)}_1, e^{(0)}_2,\dots, e^{(0)}_{N_w})={\rm Linear}({\bf x})$. Let encoder embedding ${\bf{e}}^{(N)}=(e^{(N)}_1, e^{(N)}_2,\dots, e^{(N)}_{N_w})$ be the output of $N$-layer encoder, and embedding ${\bf e}^{(i)}$ for the $i$th layer in encoder is calculated as follows:
\begin{align}
	{\bf \hat{e}}^{(i)}=&{\rm BN}\left( {\bf e}^{(i)}+	{\rm MultiHead}\left(Q_i,K_i,V_i\right) \right) \nonumber \\
	{\bf \hat{e}}^{(i+1)}=&{\rm BN}\left( {\bf \hat{e}}^{(i)}+\max\left(0, {\bf \hat{e}}^{(i)}W_1+b_1\right)W_2+b_2\right).\label{con:4}
\end{align}
where ${\rm BN}(\cdot)$ is the batch normalization operator. We use a multi-head attention module ${\rm MultiHead}(\cdot)$ with $\rm h$ parallel heads to capture the relationships between different embedded nodes, and each head ${e}^{(i)}_j$ performs a scaled dot-product attention function in eq. \eqref{con:4}. Matrix $Q_i, K_i, V_i$ are respectively the queries, keys and values for the $i$th encoder layer and the queries and keys share the same dimension $d_{k_i}$.
\begin{align}
	Q_i=W^{Q_i}{\bf{e}}^{(i)}, &K_i=W^{K_i}{\bf{e}}^{(i)},  V_i=W^{V_i}{\bf{e}}^{(i)} \nonumber \\
	{m}^{(i)}_j=&{\rm softmax}\left(\frac{Q_iK_i^T}{\sqrt{d_{k_i}}} \right)V_i \\
	{\rm MultiHead}\left(Q_i,K_i,V_i\right)=&{\rm Concat}\left( { m}^{(i)}_1, { m}^{(i)}_2, \dots {m}^{(i)}_{\rm h}\right)W^o . \nonumber
\end{align}

\subsection{Robot Decoder}
Instead of encoding all the embedding features, the encoder should give more attention to the robot related features and output the probabilities of limited options: selecting a robot. So we construct a robot mask operator ${\rm M}^{\mathcal R}(\cdot)$ related to the index of node in ${\bf{e}}^{(N)}$. It can distinguish the relevant and irrelevant embedding nodes of the robot by setting different mask values for different node index:
\begin{equation}
	{\rm M}^{\mathcal R}(l) = 
	\begin{cases}
		0, & \text{if $1\leq l\leq N_a$ \text{and} $l\in N^+$}; \\
		1, & \text{if $N_a<l\leq N_w $ \text{and} $l\in N^+$}.
	\end{cases}
\end{equation}
where $l$ is node index, which can never been changed for each embedding node $e^{(i)}_l$ during all the  processes of robot net.
In the attention operation, Let $Q_{d_r}=W^{Q_{d_r}}\cdot {\rm Concat}\left({\bf{e}}^{(N),r}, {\bf{e}}^{(N),g}\right)$ be the query calculated by a robot temporal embedding layer, where ${\bf{e}}^{(N),r}={\rm LSTM}\left(e^{(N)}_{h^a_{j-1}}, e^{(N)}_{h^a_{j-2}}, \dots\right)$  is the robot temporal embedding calculated with a LSTM layer to extract the robot sequence temporal features, and ${\bf{e}}^{(N),g}=\sum_{j=1}^{N}{e^{(N)}_j}/N_w$ is the global encoder embedding containing global static features. Key $K_{d_r}$ and value $V_{d_r}$ are computed only via ${\bf{e}}^{(N),r}$. Through a multi-head attention network and a linear layer, a new robot decoder query $\hat{Q}_{d_r}$ is generated:
\begin{align}
	\hat{Q}_{d_r}=W^{\hat{Q}_{d_r}}\cdot {\rm MultiHead}\left(Q_{d_r}, K_{d_r}, V_{d_r}\right).
\end{align}

To obtain the robot selecting probability, we construct a robot policy module composed of a dot-product attention layer with decoder mask  ${\rm M}^{\mathcal R}(\cdot)$, a tanh activation layer and a softmax layer.
First we construct the robot decoder key based on the encoder embedding ${\bf{e}}^{(N)}$:
\begin{align}
	\hat{K}_{d_r} = W^{\hat{K}_{d_r}}{\bf{e}}^{(N)} 
\end{align}

The masked compatibility can be calculated with robot decoder query $\hat{Q}_{d_r}$ and the robot decoder key $\hat{K}_{d_r}$ using a dot-product attention layer:
\begin{equation}
	u^{d_r}_j = 
	\begin{cases}
		{\rm tanh}\left(\frac{\hat{Q}_{d_r}\hat{K}_{d_r}^T}{\sqrt{d_k}}\right), & \text{if ${\rm M}^{\mathcal R}(j)=0$}; \\
		-\infty, & \text{if ${\rm M}^{\mathcal R}(j)=1$}.
	\end{cases}
\end{equation}

Based on the masked compatibility, the output robot selecting probability is calculated via a softmax layer. It is worth noting that only the probability of robot related node has positive number because ${\rm e}^{-\infty}=0$. The probability of selecting the $l$th robot equals:
\begin{align}
	p^{\theta_r}_l={\rm softmax}\left(u^{d_r}_1, u^{d_r}_2, \dots, u^{d_r}_{N_w}\right)=\frac{{\rm e}^{{u^{d_r}_l}}}{\sum_{l=1}^{N_w}{{\rm e}^{{u^{d_r}_l}}}}.
\end{align}

\subsection{Node Decoder} 
Different from robot decoder and encoder, the decoder in the node net cares not only the spatial node features, but also the temporal arc features with cycle constraints. We design a cyclic node temporal embedding layer composed of $N_{r_l}=N_g$ number of gate recurrent unit (GRU) neuron modules to obtain the cyclic temporal embedding from with the encoder embedding at $t_k$ as shown in the middle of Fig. \ref{network}.

The history  selected robots and nodes are respectively stored in the historical data pools, based on which, we can obtain the previous standard cycle number of selected node IDs ${h}_{\max{(0,k-N_{r_l}+1)}}^{r_l}, h_{\max{(0,k-N_{r_l}+2)}}^{r_l}, \dots, h_k^{r_l}$ for the $l$th robot that has just been chosen based on the robot selecting probability distribution of robot net output. Latent embedding are obtained by sequentially inputting
encoder embedding with same index of historical selected node into gate recurrent units:
\begin{equation}
\begin{split}
{ z}^{r_l}_{k+1}&={\rm GRU}\left( e_{{ h}_{k}^{r_l}}^{(N)}, {\rm GRU}\left(e_{{ h}^{r_l}_{k-1}}^{(N)}, {\rm GRU}\left(\dots,\right.\right.\right.\\ 
&\hspace{10pt}\left.\left.{\rm GRU}\left(e_{{ h}_{\max{(0,k-N_{r_l}+1)}}^{r_l}}^{(N)}, e_{{ h}_{\max{(0,k-N_{r_l}+2)}}^{r_l}}^{(N)} \right)\right)\right)
\end{split}
\end{equation}

To combine the feature of the related robot, the embedding of selected robot and embedding ${ z}^{r_l}_{k+1}$ are concatenated to obtain the node temporal embedding ${\bf e}^{\mathcal T}$:

\begin{equation}
	\begin{split}
		&{\bf e}^{\mathcal T}={\rm Concat}\left(e_{{ h}_{j}^a}^{(N)}, {z}^{r_l}_{k+1}\right).
	\end{split}
\end{equation}	

Then we further concatenate the node temporal embedding with the robot node $e^{N}_{l}$ and the global encoder embedding:
\begin{equation}
	{\bf \hat{e}}^{\mathcal T}={\rm Concat}\left(e^{N}_{l}, {\bf e}^{\mathcal T}, {\bf{e}}^{(N),g} \right).
\end{equation}

Let  ${Q}_{d_n}$ be node query that can be calculated based on embedding ${\bf \hat{e}}^{\mathcal T}$ that contains both spatial, temporal node features and the relative robot features. The key and value for multi-head attention module consider all the encoder embedding nodes, and the relative functions to compute the node decoder embedding  $\hat{\bf e}_{d_n}$ are as follows:
\begin{align}
	{Q}_{d_n}=W^{{Q}_{d_n}} { \hat{\bf e}}^{\mathcal T}, &
	{K}_{d_n}=W^{{K}_{d_n}} {\bf {e}}^{(N)},
	{V}_{d_n}=W^{{V}_{d_n}} {\bf{e}}^{(N)} \nonumber \\
	\hat{\bf e}_{d_n} =& {\rm MultiHead}\left({Q}_{d_n}, {K}_{d_n}, {V}_{d_n}\right).
\end{align}

Similar to the robot decoder, we continue to calculate the compatibility for each node:
\begin{equation}
	u^{d_n}_{j,k} = 
	\begin{cases}
		{\rm tanh}\left(\frac{\hat{Q}_{d_n}\hat{K}_{d_n}^T}{\sqrt{d_n}}\right), & \text{if ${\rm M}^{\mathcal N}(j,k)=0$}; \\
		-\infty, & \text{if ${\rm M}^{\mathcal N}(j,k)=1$}.
	\end{cases}
\end{equation}
where the new node decoder query $\hat{Q}_{d_n}=W^{\hat{Q}_{d_n}}{ \hat{\bf e}}^{d_n}$ and the new node decoder key $\hat{K}_{d_n}=W^{\hat{K}_{d_n}}{ {\bf e}}^{(N)}$ are both processed by a linear transform, respectively from the node decoder embedding and encoder embedding.
${\rm M}^{\mathcal N}(\cdot)$ is the graph node decoder mask operator to control valid action space in the node net. The graph node mask considers only the graph nodes with the index $j>N_a$, where $N_a$ is the number of robots, and it follows the arc constraints defined in robot temporal diagraph $\mathcal{G}^{r_l}_k$ for the $l$th robot at the timestamp $t=t_k^{r_l}$, which means:
\begin{equation}
	{\rm M}^{\mathcal N}(j,k) = 
	\begin{cases}
		1, & \text{if }0\leq j\leq N_a;\\
		0, & \text{elif $\mathbb{I}(v^{r_l}_{j},t^{r_l}_k)=1, \phi(h^{r_l}_{k+1}) =\hat{\sigma}_{r_l}(\phi(h^{r_l}_{k}))$}; \\
		1, & \text{otherwise}.
	\end{cases}
\end{equation}

Finally, the node net outputs probability of selecting a graph node $v^{r_l}_{j}$ via inputting compatibility into a softmax layer:
\begin{align}
	p^{\theta_n}_{j,k}={\rm softmax}\left(u^{d_n}_{1,k}, u^{d_n}_{2,k}, \dots, u^{d_n}_{N_w,k}\right)=\frac{{\rm e}^{{u^{d_n}_{j,k}}}}{\sum_{j=1}^{N_w}{{\rm e}^{{u^{d_n}_{j,k}}}}}.
\end{align}

\section{Training Method}
\label{sec:train}
In this section, we present the methods to train policies on HTAN. The training methods mainly involve a hierarchical reinforcement learning method HRC-REINFORCE and a multi-stage curriculum learning method HCR2C aiming to improve the policy generalization performance.

\subsection{Hierarchical Reinforcement Learning with Counterfactual Rollout Baseline}
REINFORCE with rollout baseline algorithm in the work of \cite{koolattention} is chosen as our basic reinforcement learning framework because other more popular methods with actor-critic structure have been proved in  \cite{koolattention} to be more difficult to learn stably in similar but simpler optimization tasks due to the challenge in designing an effective critic network. The policy gradient \cite{sutton2018reinforcement} for REINFORCE with rollout baseline based on the option model \cite{sutton1999between} can be written as:

\begin{align}
	\nabla {\mathcal J}\left({\theta}\right)&\propto {\mathbb E}_{\pi}\left[{\left(Q_{\pi}\left({s_t},{\boldsymbol o}_t\right)-b({ s}_t)\right)}\nabla_{\theta}\ln \pi\left({\boldsymbol o}_t|{s}_t\right)\right].
\end{align}
% $\mu$ is the on-policy distribution of $\pi$, and  
where $\pi$ is the policy of net with parameters $\theta$ and $b(s)$ is the rollout baseline.
 Based on that, we find the final epoch reward is related with both of the robot network policy and the graph node network policy. Such global sharing of the reward will bring about the inaccurate evaluation for policies between different policy layers, because it is difficult to distinguish which network policy causes the change of the reward, and this will lead to the poor effect of policy learning. 
We noticed that this problem in hierarchical reinforcement learning was consistent with the problem in multi-agent reinforcement learning, and the work in \cite{foerster2018counterfactual} has solved this by proposing an advantage function with counterfactual baseline in eq. \eqref{con:19}.
\begin{align}
	A^a(s,{\boldsymbol u})=Q\left(s,{\boldsymbol u}\right)-\sum_{{u'}^{a}}\pi^a\left({u'}^{a}|\tau^a\right)Q\left(s,\left({\boldsymbol u}^{-a},{u'}^{a}\right)\right). \label{con:19}
\end{align}
where $\tau^a$ is the historical observation-action sequence.  ${\boldsymbol{u}}$ is the combined actions of all agents. $Q\left(s,{\boldsymbol u}\right)$ is the action-value function. ${\boldsymbol{u}}^{-a}$ represents all other agents of joint action except agent $-a$, and ${u'}^{a}$ represents the default action for agent $a$. 

Inspired of it, we propose the hierarchical REINFORCE with counterfactual rollout baseline (HCR-REINFORCE) where each hierarchical net is regarded as an agent, and a counterfactual rollout baseline is used to evaluate the value contribution of each layer
to reduce the unfair credit assignment between hierarchical policies.
Because there are countless possibilities for each layer of output policy, to simplify the calculation, we just define one default policy for each evaluation baseline, named robot baseline and graph node baseline, respectively. The two advantage functions in HCR-REINFORCE for HTAN-robot net and HTAN-node net are defined in eq. \eqref{con:20}.
\begin{align}
	A^r(s,{\boldsymbol o})=&Q\left(s,{\boldsymbol o}\right)-b^r\left(s,(o^{-r},{o'}^r,o^g)|\theta^g_{BL}\right)\nonumber \\
	A^g(s,{\boldsymbol o})=&Q\left(s,{\boldsymbol o}\right)-b^g\left(s,(o^{r},{o}^{-g},{o'}^g)|\theta^r_{BL}\right). \label{con:20}
\end{align}
where $\boldsymbol{o}=(o^r,o^g)$ is the joint option composed of a robot option and a graph node option in sequence. ${o'}^r$ is the robot option generated by a default robot selecting heuristic algorithm.  $o^g$ is the graph node option generated by a default graph node selecting heuristic algorithm. $\theta^r_{BL}$ and $\theta^g_{BL}$ are respectively the parameters of the robot baseline net and the graph node baseline net copied from the robot net and graph node net with a lower updating frequency.
$Q\left(s,\boldsymbol{o}\right)$ is an approximate option-value function based on the rollout sampled from the joint policy of robot net and graph node net in sequence:
\begin{align}
	Q\left(s,\boldsymbol{o}\right) \doteq & {\mathbb E}_{\pi^h}{\left[G_t|S_t=s,\boldsymbol{O}_t=\boldsymbol{o}\right]} \nonumber\\
	=&{\mathbb E}_{\pi^h}{\left[\sum_{m=0}^{\max_l{(T_R^{r_l})}-t-1}{\gamma^mR_{t+m+1}|S_t=s,\boldsymbol{O}_t=\boldsymbol{o}}\right]}.
\end{align}
where $R$ is the instance reward defined in eq. \eqref{con:3}. $\pi^h$ is the joint policy of $\pi^r$ and $\pi^g$, where $\pi^r$ is the policy with parameter $\theta^r$ and $\pi^g$ is the policy with parameter $\theta^g$. $G_t = \sum_{m=0}^{\max_l{(T_R^{r_l})}-t-1}{\gamma^mR_{t+m+1}}$ is the discounted cumulative reward at the $t$th step.
The reinforcement learning loss function for HCR-REINFORCE is shown in eq. \eqref{con:22}, and algorithmic pseudocode can be referred to Algorithm \ref{alg:1} (Line 25).
\begin{align}
	&{\mathcal L}_{RL}\left({\theta_r,\theta_g}\right)\propto\mathbb{E}\left[A(s,{\boldsymbol o})
	\ln \pi(o_t|s_t,\theta_r,\theta_g)\right] \nonumber \\
	=&\mathbb{E}\left[A^r(s,{\boldsymbol o})
	\ln \pi(o_t|s_t,\theta_r)  + 
	A^g(s,{\boldsymbol o})\ln \pi(o_t|s_t,\theta_g)\right] \nonumber \\
	\doteq& -\frac{1}{B}\sum_{i=1}^{B}\left[\gamma^m \left(G_i-b_i^r\left(s,(o^{-r},{o'}^r,o^g)|\theta^g_{BL}\right)\right)\ln \pi_i(o_m|s_m,\theta_r) \right.\nonumber \\
	&+\left.\gamma^m \left(G_i-b_i^g\left(s,(o^{r},{o}^{-g},{o'}^g)|\theta^r_{BL}\right)\right)\ln \pi_i(o_m|s_m,\theta_g) \right]
	\label{con:22}
\end{align}
where $B$ is the number of instances in a batch, and $i$ is the index of instance.  $\gamma$ is the discount factor within $[0,1]$, and we set it as $\gamma=1$ here because what we care about is how to maximize the expect of epoch reward (makespan). 
HCR-REINFORCE uses Monte Carlo option-value function to get better convergence assurance, whose updating procedures can only be executed 
after an epoch ends and the trajectory data based on policy $\pi(\cdot|\cdot,\theta_r)$ and $\pi(\cdot|\cdot,\theta_g)$ is collected.

\begin{algorithm}[H]
	\caption{Hierarchy REINFORCE with Counterfactual Rollout Baseline (HCR-REINFORCE)}
	\label{alg:1}
	\renewcommand{\algorithmicrequire}{\textbf{Input:}}
	\begin{algorithmic}[1]
		\REQUIRE number of epochs $E$, batch size $B$, significance $\alpha_s$, robot net parameters $\theta_{r}$, graph node net parameters $\theta_{g}$, robot net baseline parameters $\theta^{r}_{BL}$, graph node baseline net parameters $\theta^{g}_{BL}$ %%input
		\FOR{epoch k = $1,\dots, E$}
			\STATE $s_{t_0} \leftarrow \text{RandomInstance}(B)$
			\STATE Initialize parameters and let $m=0$
			\WHILE{NotFinish($s_{t^g_m}$)}
				\STATE $o^r_m \leftarrow \text{RobotRandomSample}( \pi(o^r_m|s_{t^r_m}, \theta_r))$
				\STATE ${\hat o}^{r}_m \leftarrow {\text{CostGreedyRobot}}({\hat \pi}({\hat o}^{r}_m|s_{t^r_m}))$
				\STATE Update sequence $\mathcal{H}^a_j \leftarrow h^a_j$ based on $o^r_m$
				\STATE Update state $s_{t^g_m} \leftarrow \mathcal{H}^a_j$
				\STATE $o^g_m\leftarrow \text{GraphNodeRandomSample}(\pi(o^g_m|s_{t^g_m}, \theta_g)$
				\STATE ${\hat o}^{g}_m\leftarrow {\text{CostGreedyGraghNode}}({\hat \pi}({\hat o}^{g}_m|s_{t^g_m}))$
				\STATE ${\hat {\mathcal L}}_{r}(s_{t^g_m}, \theta_r)\leftarrow{\hat {\mathcal L}}_{r}(s_{t^g_m}, \theta_r)-C_r\cdot\ln \pi({\hat o}^{r}_m|s^g_{t_m}, \theta_r)$
				\STATE ${\hat {\mathcal L}}_{g}(s^g_{t_m}, \theta_g)\leftarrow{\hat {\mathcal L}}_{g}(s^g_{t_m}, \theta_g)-C_g\cdot\ln \pi({\hat o}^{g}_m|s^g_{t_m}, \theta_g)$
				\STATE Update reward and state $R_{t^g_{m+1}}, s_{t^r_{m+1}} \leftarrow s_{t^g_m}, o^r_m, o^g_m$
				\STATE $m \leftarrow m+1$
			\ENDWHILE
			\STATE Record final step $T = m$
			\STATE Stack trajectory $s_{t^r_0}, o^r_0, s_{t^g_0}, o^g_0, R_{t^g_0}, s_{t^r_1}, o^r_1, s_{t^g_1}, o^g_1, R_{t^g_1}$, $\cdots$, $s_{t^r_T}, o^r_T, s_{t^g_T}, o^g_T, R^g_{t_T}$
			\FOR{$m = 0,\dots, T-1$}
				\STATE  $G \leftarrow \sum_{t=m+1}^{T}\gamma^{t-m-1}R_{t}$
				\STATE ${\hat {\mathcal L}}_{ BC}(\theta_r,\theta_g)\leftarrow{{\hat {\mathcal L}}_{r}(s_{t_m}, \theta_r)}+{\hat {\mathcal L}}_{g}(s_{t_m}, \theta_g)$			%Calculate behavior cloning loss 
				\STATE  $b^r\left(s,(o^{-r},{o'}^r,o^g)|\theta^g_{BL}\right)$ %Calculate the counterfactual baseline for robot net 
				\STATE 
				$b^g\left(s,(o^r,o^{-g},{o'}^g)|\theta^r_{BL}\right)$ %Calculate the counterfactual baseline for graph node net  
				\STATE $A^r(s,{\boldsymbol o})\leftarrow G-b^r\left(s,(o^{-r},{o'}^r,o^g)|\theta^g_{BL}\right)$
				\STATE $A^g(s,{\boldsymbol o})\leftarrow G-b^g\left(s,(o^{r},{o}^{-g},{o'}^g)|\theta^r_{BL}\right)$
				\STATE  ${\mathcal{L}}_{RL}(\theta_r,\theta_g) \leftarrow-\frac{1}{B}\sum_{i=1}^{B}\left[\gamma^m A_i^r(s,{\boldsymbol o})\ln \pi_i(o_m|s_{t_m},\right.$ $\left.\theta_r)+\gamma^m A_i^g(s,{\boldsymbol o})\ln \pi_i(o_m|s_{t_m},\theta_g) \right]$ %Calculate the reinforcement loss
				\STATE $L(\theta_r,\theta_g)$$\leftarrow$$\eta^k{\hat {\mathcal L}}_{ BL}(\theta_r,\theta_g)+ (1-\eta^k){\mathcal L}_{RL}(\theta_r,\theta_g)$ %Calculate hybrid loss 
				\STATE $\theta_r \leftarrow \text{Adam}(\theta_r, \nabla L)$
				, $\theta_g \leftarrow \text{Adam}(\theta_g, \nabla L)$
			\ENDFOR
			\STATE $m^r \leftarrow 1/B\sum_{i=1}^{B}\left[b^g_i\left(s,({o}^r,o^g)|\theta^r_{BL},\theta^g\right)\right]$
			\STATE $m^g \leftarrow 1/B\sum_{i=1}^{B}\left[b^r_i\left(s,({o}^r,o^g)|\theta^r,\theta^g_{BL}\right)\right]$
			\STATE $m \leftarrow 1/B\sum_{i=1}^{B}\left[b_i\left(s,({o}^r,o^g)|\theta^r,\theta^g\right)\right]$
			\IF {$m<m^r$}
			\IF {$\text{OneSidedPariedTTest}\left(m,b^r\left(s,(o^{-r},{o'}^r,o^g)|\theta^g_{BL}\right)\right)$ $<\alpha_s$}
			\STATE Update robot baseline net  $\theta^{r}_{BL}\leftarrow \theta_r$
			\ENDIF
			\ENDIF			
			\IF {$m<m^g$}
			\IF {$\text{OneSidedPariedTTest}\left(m,b^g\left(s,(o^r,o^{-g},{o'}^r)|\theta^r_{BL}\right)\right)$ $<\alpha_s$}
			\STATE Update graph node baseline net  $\theta^{g}_{BL}\leftarrow \theta^g$
			\ENDIF
			\ENDIF
		\ENDFOR
	\end{algorithmic}
\end{algorithm}

The early stage of reinforcement learning is slow because it requires a lot of trial and error. 
To overcome this challenge, we use a joint optimization learning technique  to accelerate the early stage learning speed of HCR-REINFORCE by adding the behavior cloning loss at the procedure of data collecting into the loss of HCR-REINFORCE and we gradually reduce the loss weight of behavior cloning to reduce the exploratory influence on reinforcement learning as much as possible.

 To calculate the behavior cloning loss at the $t_m$th step, we sample options for each layer based on the default heuristic algorithm to get a evaluation label, and record the logit of selecting the option related to the behavior cloning label:

\begin{align}
	{\hat {\mathcal L}}_{r}(s_{t^r_m}, \theta_r)=&-C_r\cdot\ln \pi({\hat o}^{r}_m|s_{t^r_m}, \theta_r)  \nonumber \\
	{\hat {\mathcal L}}_{g}(s_{t^g_m}, \theta_g)=&-C_g\cdot\ln \pi({\hat o}^{g}_m|s_{t^g_m}, \theta_g). \label{con:23}
\end{align}
where ${\hat {\mathcal L}}_{r}(s_{t_m}, \theta_r)$ is the robot behavior cloning loss for state $s_{t^r_m}$  and robot net with net parameters $\theta_r$. ${\hat {\mathcal L}}_{g}(s_{t_m}, \theta_g)$ is the graph node behavior cloning loss for state $s_{t^g_m}$ and graph node net with parameters $\theta_g$. In order to evaluate the similarity between the action selected by the neural network and the label action, we calculate the logit $\ln \pi({\hat o}^{r}_m|s_{t_m}, \theta_r)$ and $\ln \pi({\hat o}^{g}_m|s_{t_m}, \theta_g)$ corresponding to the output label action of the neural network and scale it.

After the trajectory data required is obtained, we compute the loss function and update the hierarchical network parameters. The joint loss function is computed by eq. \eqref{con:24}, and we call the method combing the REINFORCE with counterfactual rollout baseline and behavior cloning as the joint optimization learning method.
\begin{align}
	L(\theta_r,\theta_g)=\eta^k{\hat {\mathcal L}}_{ BL}(\theta_r,\theta_g)+ (1-\eta^k){\mathcal L}_{RL}(\theta_r,\theta_g) \label{con:24}
\end{align}
where $\eta$ is an attenuation coefficient within $[0,1]$($\eta=0.9$), and ${\hat {\mathcal L}}_{ BL}(\theta_r,\theta_g)=\sum_{m=0}^{\max_l{(T_R^{r_l})}}\left[{\hat {\mathcal L}}_{r}(s_{t^r_m}, \theta_r)+{\hat {\mathcal L}}_{g}(s_{t^g_m}, \theta_g)\right]$ is the behavior cloning loss for the whole hierarchical network. ${\mathcal L}_{RL}(\theta_r,\theta_g)$ is the reinforcement learning loss in eq. \eqref{con:22}

\subsection{Multi-Stage Curriculum Learning}
To further improve the policy scaling up and generalization performance,
we design a multi-stage curriculum learning method (HCR2C) based on the work of HCR-REINFORCE. One simple approach is to provide the trainer with a greater variety of random training instances. However, directly incorporating highly randomized training data during the initial stages may result in significant variance and hinder efficient policy learning. As a solution, we have developed a multi-stage curriculum learning approach that gradually expands the range of randomness and maximum instance scale. To mitigate reward variance, we adjust a new instant reward function as $R_c=R/\alpha$ to minimize the potential disruptions caused by drastic changes in rewards, where $\alpha=N_r/{N_a}$ is a scale factor to describe the scale complexity.

The multi-stage curriculum learning process is illustrated in Algorithm \ref{alg:2} in details. 
$N$ stage of curricula  $\mathfrak{C}=\{C_1,C_2,\dots,C_N\}$ are set up, where $C_i=\{U(N_a^{-},N_a^{+}),U(N_r^-,N_r^+),U(N_s^-,N_s^+)\}$, $i=1,2,\dots,N$ contains configurations about the distribution of each variable scale parameter in the $i$th curriculum, and $U(\cdot)$ is a uniform distribution. 
As the curriculum stage increases, we systematically reduce the lower limit $N_{[\cdot]}^{-}$ and increase the upper limit $N_{[\cdot]}^+$. This design ensures that the higher-level curriculum learned later encompasses the training scope of earlier lower-level curricula, thereby mitigating the issue of catastrophic forgetting in curriculum learning.

\begin{algorithm}[H]
	\caption{Multi-Stage Curriculum Learning based on HCR-REINFORCE (HCR2C)}
	\label{alg:2}
	\renewcommand{\algorithmicrequire}{\textbf{Input:}}
	\begin{algorithmic}[1]
		\REQUIRE number of curricula $N$, robot net parameters $\theta_{r}$, graph node net parameters $\theta_{g}$, robot net baseline parameters $\theta^{r}_{BL}$, graph node baseline net parameters $\theta^{g}_{BL}$ %%input
		\FOR{i = $1,\dots, N$} 
			\STATE $N_a^{-} \leftarrow \max (1, N_a^{-}-\Delta n_a^-)$, $N_a^{+} \leftarrow N_a^{+}+\Delta n_a^+$
			\STATE $N_r^{-} \leftarrow \max (1, N_r^{-}-\Delta n_r^-)$, $N_r^{+} \leftarrow N_r^{+}+\Delta n_r^+$			
			\STATE $N_s^{-} \leftarrow \max (1, N_s^{-}-\Delta n_r^-)$, $N_s^{+} \leftarrow N_s^{+}+\Delta n_s^+$	
			\STATE generate a joint distribution for curriculum $C_i=\{U(N_a^{-},N_a^{+}),U(N_r^-,N_r^+),U(N_s^-,N_s^+)\}$
			\STATE Sample instances $s_{t_0}\sim  \text{RandomInstance}(C_i)$ 
			\STATE $\theta_{r}, \theta_{g}\leftarrow \text{HCR-REINFORCE}(s_{t_0},R_c)$
		\ENDFOR
	\end{algorithmic}
\end{algorithm}

\section{ EXPERIMENTS AND RESULTS}
\label{sec:result}
In this section, we systematically present our results through a series of experiments to showcase the performance of our method for MRTP in RMFS and the comparison experiments are conducted on both simulated and real-world instances.  
First,  we introduce the settings for instances in simulated experiments in detail.
Then, we demonstrate the training results of HCR-REINFORCE on fixed scale simulated instances and show the ablation results to analyze the impact of different enhancement elements on the ultimate performance. 
The performance is analyzed by comparing the test results of HCR-REINFORCE with various state-of-the-art methods.
For generalization, we train HCR-REINFORCE with multi-stage curricula, and the training curves with HCR2C are shown to analyze the curriculum learning results. Also, we test our planner trained with HCR2C and compare its performance with other methods on various maps and random scales that up to 200 robots. On real-world instances, we compare the performance of our planner with the native company task planner considering the indicators of both makespan and planning speed.
In order to ensure fairness, all the experiments at training stage and execution stage are conducted on the same framework introduced in Section \ref{sec:framework}  on the device with a GeForce RTX 4090 GPU and 20 GB video memory.

\subsection{Simulated Setup}
\subsubsection{On Fixed Scale Simulated Instances}

According to the parameters of the number of robots, the number of retrieval racks and the number of unoccupied storage locations, we set 16 kinds of parameter configuration schemes for fixed small-scale, medium-scale and large-scale simulated instances, and the detailed parameter values are shown in Table \ref{tab:fixed-scale setup}. This configuration scheme will be used to generate training and test instance sets for HCR-REINFORCE and other comparison algorithms. In the training stage, HCR-REINFORCE and state-of-the-art comparison methods will be trained in turn on 1024 instances randomly generated according to each scale configuration scheme, and each method finally generates 16 kinds of policy models corresponding to each fixed scale. In the execution stage, 100 test instances are randomly generated for each configuration parameter according to the parameters in Table \ref{tab:fixed-scale setup} as a test set with fixed-scale.
\begin{table}[!t]
	\caption{EXPERIMENT SETTINGS FOR FIXED-SCALE SIMULATED INSTANCES\label{tab:fixed-scale setup}}
	\centering
	\renewcommand{\arraystretch}{1.0}	
	\begin{tabular}{cccccc}
		\Xhline{1pt}
		%\multicolumn{11}{|l|}  \\
		%\hline \hline
		Scale & Instance ID & $N_a$  &  $N_r$  & $N_s$ & Batch Size \\
		\Xhline{0.7pt}		
		\multirow{8}{*}{Small-Scale} & F1     & 2     & 4         & 4 & 512           \\ 
		&F2     & 2     & 4    & 8      & 512      \\
		&F3     & 2     & 6     & 6     & 512      \\
		&F4	  & 2     & 6     & 12     & 512       \\
		&F5     & 2     & 8     & 8      & 256     \\
		&F6    & 2     & 8     & 16      & 256     \\
		&F7   & 2     & 10     & 10      & 256    \\			
		&F8   & 2     & 10     & 20      & 256    \\
		\hline	
		\multirow{6}{*}{Medium-Scale}&F9     & 5     & 10    & 10      & 128      \\
		&F10     & 5     & 10     & 20      & 128     \\
		&F11	  & 5     & 15     & 15   & 128         \\
		&F12     & 5     & 15     & 30        & 128   \\
		&F13    & 5     & 20     & 20       & 128    \\
		&F14   & 5     & 20     & 40         & 128 \\	
		\hline		
		\multirow{2}{*}{Large-Scale} & F15   & 10     & 20     & 20        & 64  \\		
		&F16     & 10     & 20         & 40    & 64       \\ 	
		\Xhline{1pt}
	\end{tabular}
\end{table}

\subsubsection{On Random Scale Simulated Instances}
\begin{table}[!t]
	\caption{EXPERIMENT SETTINGS FOR RANDOM SCALE MAPS\label{tab:random map}}
	\centering
	\renewcommand{\arraystretch}{1.1}
	\begin{tabular}{cccccc}
		\Xhline{1pt} 
		%\multicolumn{11}{|l|}  \\
		%\hline \hline
		Map ID & $N_{zc}$ & $N_{zr}$ & $\max N_a$  &   $\max N_{r}+ N_{s}$  & $\max N_p$ \\
		\Xhline{0.7pt} 		
		M1     & 2     & 2   & 5   &  90     & 4            \\ 
		M2     & 5     & 2   & 10   &  180     & 4            \\ 
		M3     & 7     & 2   & 20   &  240     & 6            \\ 
		M4     & 7     & 5   & 30   &  420     & 8         \\ 
		M5     & 7     & 7   & 40   &  560     & 8       \\ 
		M6     & 7     & 10   & 50   &  770     & 12      \\ 
		M7     & 10     & 10   & 60   &  1100     & 16        \\ 
		M8     & 15     & 15   & 200   &  2400     & 40           \\ 		
		M9     & 20     & 20   & 350   &  4200   & 40 \\ 	  
		\Xhline{1pt} 											
	\end{tabular}
\end{table}

\begin{figure}[!t]
	\centering
	\includegraphics[width=2.8in]{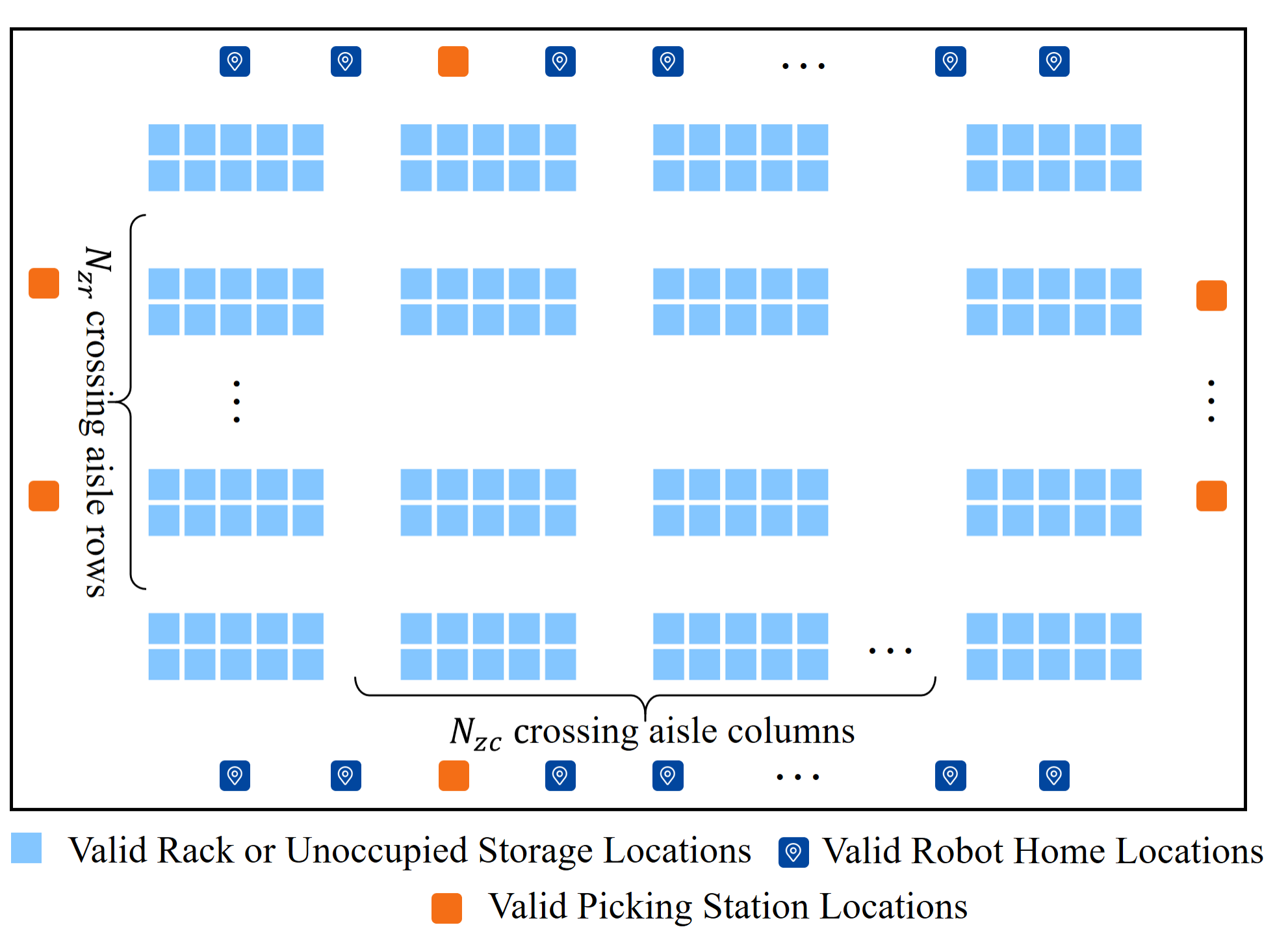}
	\caption{The diagram of map layout for M1-M9 maps.}
	\label{maplayout}
\end{figure}

\begin{table*}[!t]
	\caption{EXPERIMENT SETTINGS FOR RANDOM SCALE SIMULATED INSTANCES\label{tab:random instance}}
	\centering
	\renewcommand{\arraystretch}{1.1}
	\begin{tabular}{ccccccc}
		\Xhline{1pt} 
		Instance ID &   $N_a$     &  $N_r$  &  $N_s$ &  $N_p$   & Max $N_w$ & Map Id  \\
		\Xhline{0.7pt} 		
		U1          & $U(1, 3)$  & $U(1, 15)$   & $U(1, 30)$ &  $U(1, 4)$ &52   & M1 \\ 
		U2          & $U(1, 5)$  & $U(1, 25)$   & $U(1, 50)$  &  $U(1, 4)$  &84 & M2 \\ 
		U3          & $U(1, 10)$  & $U(1, 50)$   & $U(1, 100)$  &  $U(1, 6)$&166  & M3 \\ 	
		U4          & $U(1, 15)$  & $U(1, 75)$   & $U(1, 150)$   &  $U(1, 8)$&248 & M4 \\		
		U5          & $U(1, 20)$  & $U(1, 100)$   & $U(1, 200)$  &  $U(1, 8)$&328  & M5 \\		
		U6          & $U(1, 30)$  & $U(1, 150)$   & $U(1, 300)$  &  $U(1, 12)$&492  & M6 \\	
		U7          & $U(1, 50)$  & $U(1, 250)$   & $U(1, 500)$  &  $U(1, 16)$&816  & M7 \\		
		U8          & $U(1, 100)$  & $U(1, 500)$   & $U(1, 1000)$  &$U(1, 40)$ &1640 & M8 \\		
		U9          & $U(1, 200)$  & $U(1, 1000)$   & $U(1, 2000)$  &  $U(1, 40)$& 3240 & M9 \\						
		\Xhline{1pt} 
	\end{tabular}
\end{table*}

\begin{figure*}[!t]
	\centering
	\includegraphics[width=6.8in]{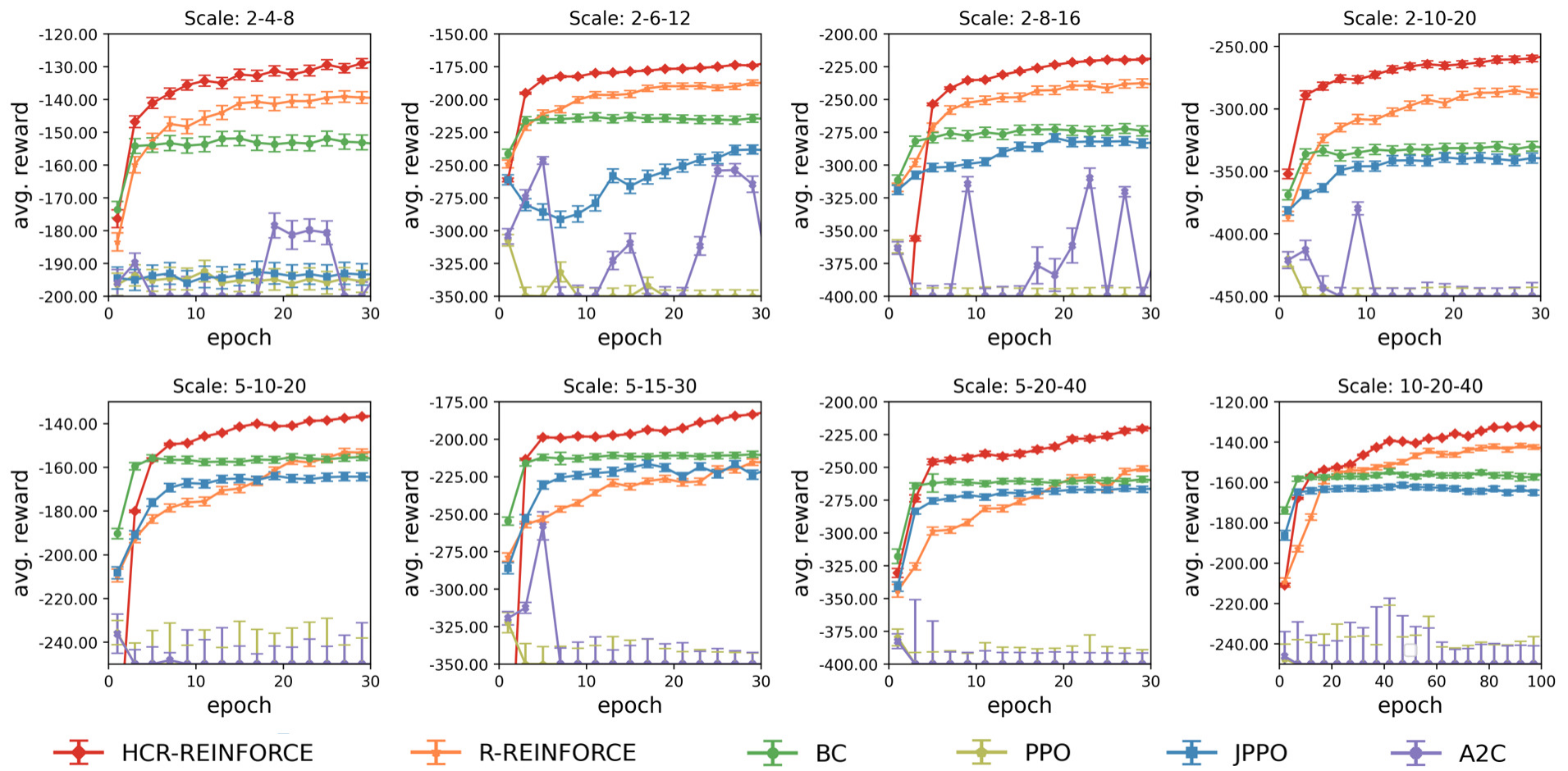}
	\caption{Comparison results of training curves for various methods on different fixed-scale simulated instances.}
	\label{fixtrain}
\end{figure*}

To enhance scaling up and generalization performance, we train a universal policy model with HCR2C to effectively address different scales of robots, racks, and unoccupied storage locations on different unlearned maps.
To evaluate the performance of HCR2C in various random scale instances and unlearned maps, we establish 9 types of instance configurations, and Table \ref{tab:random instance} presents the specific parameters for each scale configuration scheme. The random scale parameters in configuration schemes U1-U9 are generated based on maps M1-M9, whose specific map parameters are  detailed in Table \ref{tab:random map}.
As shown in Fig. \ref{maplayout}, the whole map is divided into zones by $N_{zc}$ number of aisles and $N_{zr}$ number of cross aisles. Each zone in the middle contains up to 10 locations where  retrieval racks or unoccupied storage locations can be placed. The home locations for robots and picking stations are randomly placed on the margin of the map.

In the case of U9, a maximum of 200 robots are randomly distributed on the M9 map following a uniform distribution, where 1 to 1000 retrieval  racks and 1 to 2000 initial unoccupied storage locations are also randomly generated based on the uniform distribution.

\subsubsection{Training Configuration}
The decay coefficient of HCR-REINFORCE is set as $\eta=0.99$. The robot BC loss coefficient in HCR-REINFORCE is set as $C_r=10$, and the graph node BC loss coefficient is set as $C_g=10$. The learning rate is set as $l_r=0.0001$ and the learning decay rate is $0.99$. The number of HTAN encoder layers is set as $N=2$, the dimension of encoder embedding is set as 128, the number of head in multi-head attention module is set as ${\rm h}=4$, and the training batch size of each instance is shown in Table \ref{tab:fixed-scale setup}. We uniformly use Adam as the optimizer for neural network updates.

\subsection{Training Results and Ablation Study}
\subsubsection{Comparison Results of Training Performance}

The convergence of learning curves for MRTP on fixed scale simulated instances are shown in Fig. \ref{fixtrain} to compare the tendency of average epoch reward during training process.
After training for 30 epochs on small and medium scale instances and 100 epochs on large scale instances,
it is clear that HCR-REINFORCE exhibits more rapid convergence toward higher rewards across different fixed-scale instances compared with other methods.
The state-of-the-art methods we chose to make a comparison are PPO, R-REINFORCE, JPPO, BC \cite{BC} and A2C, widely involving with different types of popular learning methods.
On small and medium scale instances, our method HCR-REINFORCE consistently achieves higher average of epoch rewards, surpassing other learning methods after 6 training rounds. 
For large-scale instances, we extend the maximum training epoch to 100 considering the slower learning speed with increasing scale. It is evident that HCR-REINFORCE still outperforms other methods in terms of final reward.
The better convergence of HCR-REINFORCE may be due to that the baseline we design can more effectively reduce variance.
By contrast, the native A2C and PPO both fail to converge due to their reliance on the critic baseline network for estimating the value function, which is a challenge exacerbated by temporal observations and constraints in our studied problem. This observation aligns with findings from previous work \cite{koolattention}. While JPPO also utilizes a critic as the baseline, its incorporation of BC loss ensures minimal deviation between action probability distribution output from the policy network and labels. Because the excessive dependence of BC loss prevents the policy performance to exceed labels, it fails to exceed our method HCR-REINFORCE.

\subsubsection{Ablation Study}
To analyze the key improvement elements of HCR-REINFORCE on the final task planning performance, we conduct ablation studies from four perspectives:  framework structure, network input,  network structure, and  training technique. To ensure fairness, we systematically remove one of the four improvement modules at a time while keeping the others unchanged, and compare their learning curves with that of  unaltered HCR-REINFORCE method.

\begin{figure*}[!t]
	\centering
	\begin{minipage}{0.23\linewidth}
		\centering
		\subfloat[framework structure]{\includegraphics[width=1.7in]{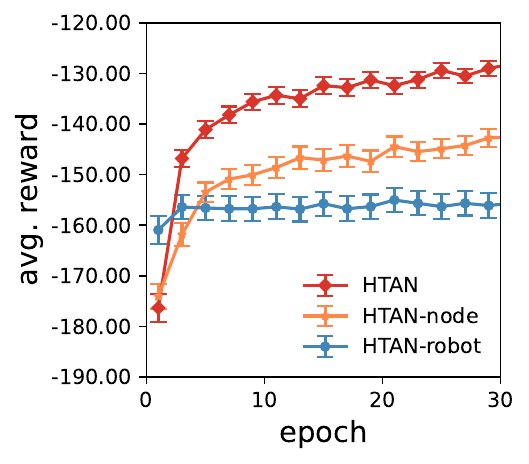}%
			\centering
			\label{framework_ablation}}
	\end{minipage}
	\begin{minipage}{0.23\linewidth}
		\centering
		\subfloat[network input]{\includegraphics[width=1.7in]{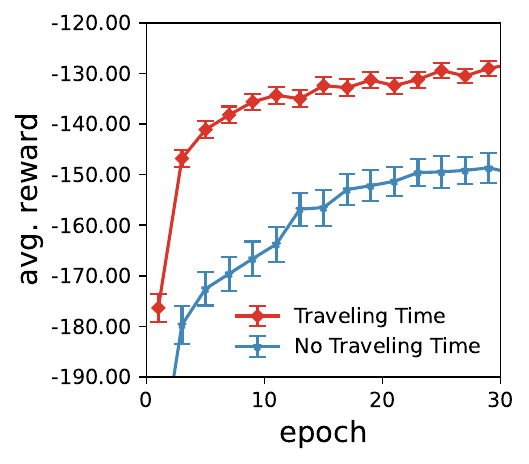}%
			\centering
			\label{network_input}}
	\end{minipage}
	\begin{minipage}{0.23\linewidth}
		\centering
		\subfloat[network structure]{\includegraphics[width=1.7in]{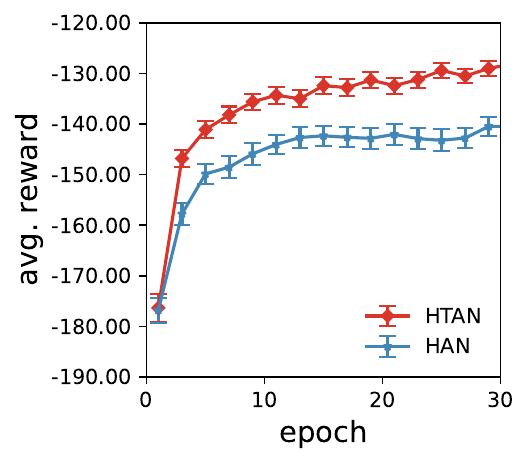}%
			\centering{
				\label{net_ablation}}}
	\end{minipage}
	\begin{minipage}{0.23\linewidth}
		\centering
		\subfloat[training technique]{\includegraphics[width=1.7in]{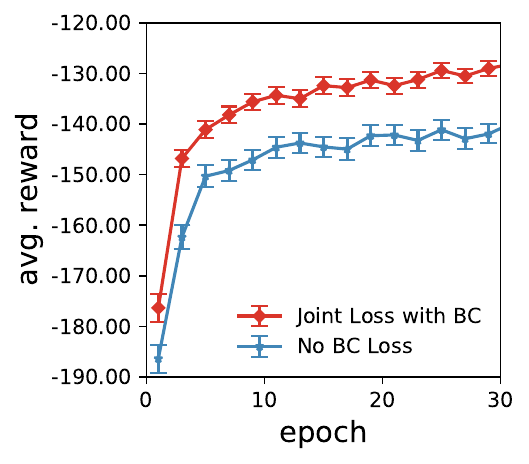}%
			\centering
			\label{hybrid_ablation}}
	\end{minipage}
	\caption{The ablation results of (a) framework structure, (b) network input, (c) network structure, and (d) training technique on the  F2 fixed-scale simulated instances.}
	\label{fig:ablation}
\end{figure*}

The results of these ablation studies are shown in Fig.  \ref{framework_ablation}-\ref{hybrid_ablation}, in which, Fig. \ref{framework_ablation} shows the comparison results of the training curves that respectively with complete HTAN structure, single HTAN-robot, and HTAN-node network. When the framework contains only single policy layer, the other layer does not employ neural network policy but is replaced with the heuristic method STNN. Fig. \ref{framework_ablation} shows that HCR-REINFORCE with complete hierarchical network structure is able to learn better policies with higher epoch rewards after training 4 epochs.

Fig. \ref{network_input} presents the results of whether input vector in HTAN contains  shared robot traveling time $r_{k,l}$ is beneficial for training performance or not. It is obvious that  the curve related to input vector containing shared robot traveling time can achieve a higher epoch reward and has faster convergence speed, and this result suggests that robot traveling time is an important factor for the performance of the planner, possibly because it is a critic representation of historical information. 

Fig. \ref{net_ablation} illustrates the results of the ablation study on network structure, comparing the training results of networks with and without the temporal embedding layers in decoder of HTAN. The network that removed temporal embedding layers in both HTAN-robot and HTAN-node is called hierarchical attention network (HAN). It is evident that our network HTAN with temporal embedding layers achieve higher epoch rewards and faster convergence compared with HAN, primarily attributed to their better ability to extract temporal features in dynamic cycle constraint and varying graph nodes.

Fig. \ref{hybrid_ablation} shows the results of  learning curves with different training techniques. The red curve shows the training result of HCR-REINFORCE  with a joint optimization training technique that mixing the loss of reinforcement learning with that of BC. On the contrary, the blue curve is the result of HCR-REINFORCE without joint optimization training technique.
The results demonstrate that HCR-REINFORCE with joint optimization training technique has a faster initial rise and ultimately achieves higher rewards. This is primarily attributed to the significant weight of BC in the early training stage, which mitigates ineffective exploration and facilitates the acquisition of positive experiences. As the training progresses, the impact on exploration ability from BC diminishes due to its decreasing weight, allowing for less interference with the reinforcement learning algorithm. Consequently, within the same training duration, the training method with joint optimization training technique attains a higher reward value.

\subsubsection{Comparison Results of Test Performance}
To objectively analyze the test performance of HCR-REINFORCE, we conduct a comparison experiment between 16 trained models with HCR-REINFORCE and start-of-the-art learning or heuristic methods on 100 test instances with configuration schemes of F1-F16. We evaluate both the quality and speed performance of the method. The quality indicator is measured by average value of makespan $\mathfrak{V}$. We use HCR-REINFORCE as the benchmark to compare the gap of $\mathfrak{V}$ between various methods:
\begin{align}
	\text{Gap}_i= \frac{\mathfrak{V}_i-\mathfrak{V}_{\text{HCR-REINFORCE}}}{\mathfrak{V}_{\text{HCR-REINFORCE}}}\times 100\%. \label{con:25}
\end{align}

Considering there is no heuristic method that can be fully adapted to our hierarchical temporal framework so far, we propose a shortest traveling time and nearest neighbor (STNN)  method by making some small improvements on the basis of existing methods of shortest leg (SL) \cite{bibtex3} and nearest neighbor (NN) \cite{NN} to adapt to our planning framework:
in the robot selecting layer, the robot with the shortest traveling time at current decision step will be chosen, and then in the graph node selecting layer, the closest valid location node will be assigned for the selected robot. 
Also, we adjust NN to adapt the hierarchical framework: the valid pair of location node and corresponding robot with the smallest evaluated traveling time from the current location of each robot to their closest node will be chosen as the planning result for next step.

\begin{figure*}[!t]
	\centering
	\includegraphics[scale=0.62]{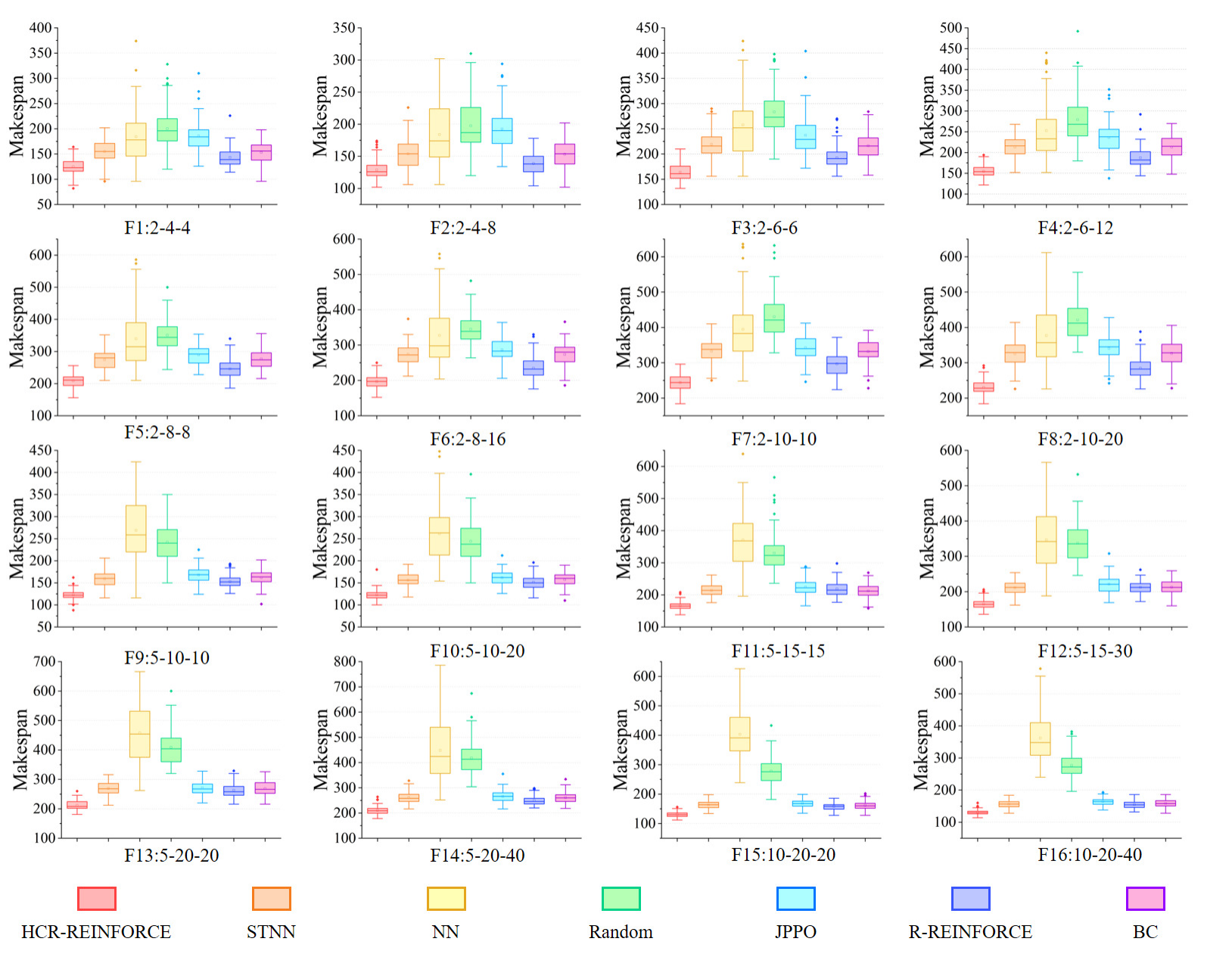}
	\caption{Box plot of test distribution results of makespan indicator on F1-F16 fixed-scale simulated instances.}
	\label{fixtest}
\end{figure*}

\begin{table*}[!t]
	\caption{RESULTS OF MAKESPAN INDICATOR ON F1-F16 FIXED-SCALE SIMULATED INSTANCES.
		\label{tab:fixtest}}
	\renewcommand{\arraystretch}{1.1}	
	\centering
	\resizebox{17.8cm}{!}{
		\begin{tabular}{c c cc cc cc cc cc cc}
			\Xhline{1pt}
			{Instance}
			& \multicolumn{1}{c}{{HCR-REINFORCE}} & \multicolumn{2}{c}{{JPPO}} &  \multicolumn{2}{c}{{R-REINFORCE}} &  \multicolumn{2}{c}{{BC}} & \multicolumn{2}{c}{{STNN}} & \multicolumn{2}{c}{{NN}} & \multicolumn{2}{c}{{Random}} \\
			%\cline{3-4} \cline{4-5}
			\cmidrule(lr){2-2} 
			\cmidrule(lr){3-4} \cmidrule(lr){5-6}  \cmidrule(lr){7-8}  \cmidrule(lr){9-10} \cmidrule(lr){11-12} \cmidrule(lr){13-14} 
			{ID}&$\mathfrak{V}(s)$ &$\mathfrak{V}(s)$ &Gap & $\mathfrak{V}(s)$ &Gap & $\mathfrak{V}(s)$ &Gap & $\mathfrak{V}(s)$ &Gap & $\mathfrak{V}(s)$ &Gap & $\mathfrak{V}(s)$ &Gap   \\
			\Xhline{0.7pt}
			{F1}&	{124.68}     &185.70 	&48.94\%	&143.26 	&14.90\%	&153.12 	&22.81\%	&154.78 	&24.14\%	&184.08 	&47.64\%	&200.28 	&60.64\%	  \\
			{F2}&	{128.72} 	&192.60 	&49.63\%	&138.68 	&7.74\%	    &153.16 	&18.99\% 	&153.84 	&19.52\%	&183.88 	&42.85\%	&197.46 	&53.40\%  \\
			{F3}&	{164.12} 	&237.10 	&44.47\%	&192.58 	&17.34\%	&216.34 	&31.82\%  	&219.28 	&33.61\%	&257.74 	&57.04\%	&283.40 	&72.68\%\\
			{F4}&	{154.58}	    &235.82 	&52.56\%	&187.30 	&21.17\%	&213.08 	&37.84\%    &212.88 	&37.72\%	&252.22 	&63.16\%	&279.12 	&80.57\%\\
			{F5}&	{208.78} 	&289.60 	&38.71\%	&245.80 	&17.73\%	&276.58 	&32.47\%    &274.02 	&31.25\%	&339.42 	&62.57\%	&351.06 	&68.15\% \\
			{F6}&	{196.98} 	&286.64 	&45.52\%	&234.62 	&19.11\%	&273.36 	&38.78\%    &273.68 	&38.94\%	&327.26 	&66.14\%	&345.00 	&75.14\%  \\
			{F7}&	{243.80} 	&342.48 	&40.48\%	&296.60 	&21.66\%	&331.76 	&36.08\% 	&332.60 	&36.42\%	&394.50 	&61.81\%	&429.86 	&76.32\%  \\
			{F8}&	{230.10} 	&340.66 	&48.05\%	&284.46 	&23.62\%	&326.34 	&41.83\%	&325.12 	&41.30\%	&376.64 	&63.69\%	&420.48 	&82.74\%   \\
			{F9}&	{122.77} 	&167.69 	&36.59\%	&152.72 	&24.40\%	&161.42 	&31.48\% 	&158.85 	&29.39\%	&268.62 	&118.80\%	&241.96 	&97.08\%  \\
			{F1}0&	{122.69}	    &161.60 	&31.71\%	&151.20 	&23.24\%	&156.98 	&27.95\%   & 156.02 	&27.17\%	&261.02 	&112.75\%	&244.16 	&99.01\% \\
			{F1}1&	{165.49} 	&223.86 	&35.27\%	&218.65 	&32.12\%	&212.01 	&28.11\%   & 213.74 	&29.16\%	&370.62 	&123.95\%	&330.07 	&99.45\% \\
			{F12}&	{164.88} 	&221.15 	&34.13\%	&212.21 	&28.71\%	&212.75 	&29.03\%   & 210.76 	&27.83\%	&346.92 	&110.41\%	&336.93 	&104.35\%  \\
			{F13}&	{211.87} 	&271.09 	&27.95\%	&263.16 	&24.21\%	&268.48 	&26.72\%   & 269.63 	&27.26\%	&457.49 	&115.93\%	&409.24 	&93.16\%    \\
			{F14}&	{210.32} 	&265.39 	&26.18\%	&249.49 	&18.62\%	&260.85 	&24.03\%   & 261.21 	&24.20\%	&447.86 	&112.94\%	&417.68 	&98.59\%    \\
			{F15}&	{130.46} 	&168.22 	&28.94\%	&158.01 	&21.12\%	&161.00 	&23.41\%   & 162.87 	&24.84\%	&402.93 	&208.85\%	&278.26 	&113.29\%   \\
			{F16}&	{130.53}     &163.54 	&25.29\%	&154.65 	&18.48\%	&158.31 	&21.28\%   & 156.12 	&19.60\%	&361.72 	&177.12\%	&277.03 	&112.23\%	   \\
			%\hline
			\Xhline{0.7pt}
			\textbf{AVG.}     &			& 			&38.40\%	&				&{20.88\%}	&			&29.54\%          &	        &29.52\%    &           &96.60\%	&   		&86.67\% \\
			\Xhline{1pt}
	\end{tabular}}
\end{table*}

\begin{figure}[!t]
	\centering
	\includegraphics[scale=0.66]{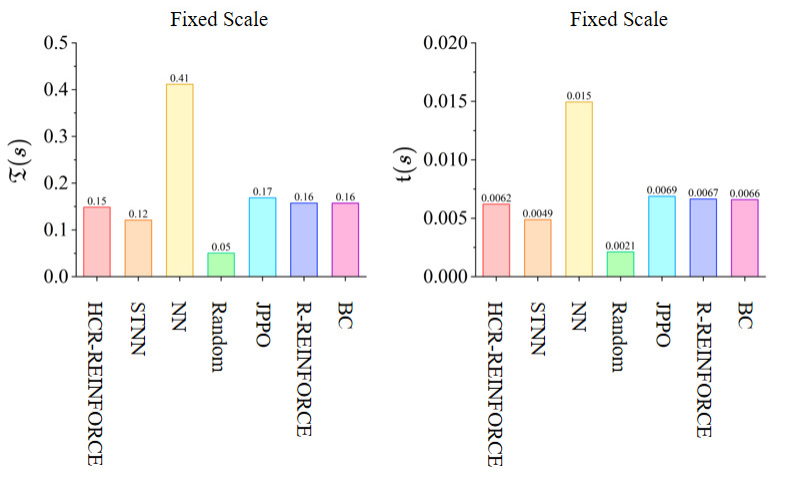}
	\caption{Comparison results of running time indicator with different methods on fixed-scale simulated instances.}
	\label{wholet}
\end{figure}

The test results of makespan indicator with different methods are presented in Table \ref{tab:fixtest}. HCR-REINFORCE outperforms other comparison methods for each instance. Although R-REINFORCE shows the best performance, there is still an average gap of $20.88\%$ between it and our HCR-REINFORCE. To further analyze the stability performance, Fig. \ref{fixtest} utilizes a box plot to display the distribution of makespan indicator solved by various methods in F1-F16 fixed-scale simulated instances. As depicted in Fig. \ref{fixtest}, the stability of NN deteriorates with increasing scale size, with even the mean value of makespan indicator surpassing that of Random method in medium and large scale instances after F5. This may be attributed to uneven task distribution of robots when greedily choosing only the nearest strategy without considering historical robot traveling time information, leading to increased total task completion time on the large scale instances. In contrast, STNN and learning methods exhibit greater stability. Among them, HCR-REINFORCE demonstrates superior performance with lower mean, median and variance values compared to other state-of-the-art methods across different test instances.
Furthermore, the speed of planning is a crucial factor that impacts the practical application of task planning methods. Fig. \ref{wholet} presents an analysis of the running speed performance for different methods with the bar chart, indicating the average total time for solving all tasks $\mathfrak{T}$ and the average time for generating one-step joint option $\mathfrak{t}$ across 100 test instances. It is evident that the random algorithm exhibits the fastest performance, followed by STNN, which aligns with expectations given their respective algorithm complexities of $O(1)$ and $O(n)$. HCR-REINFORCE demonstrates only slightly slower results than STNN across all test instances, with a mere $0.0012s$ difference in single-step union option generation time compared to STNN in milliseconds on the largest F16 instance.

\subsection{Generalization Study}

\subsubsection{Curriculum Learning Results}

\begin{figure}[!thb]
	\centering
	\includegraphics[width=3.5in]{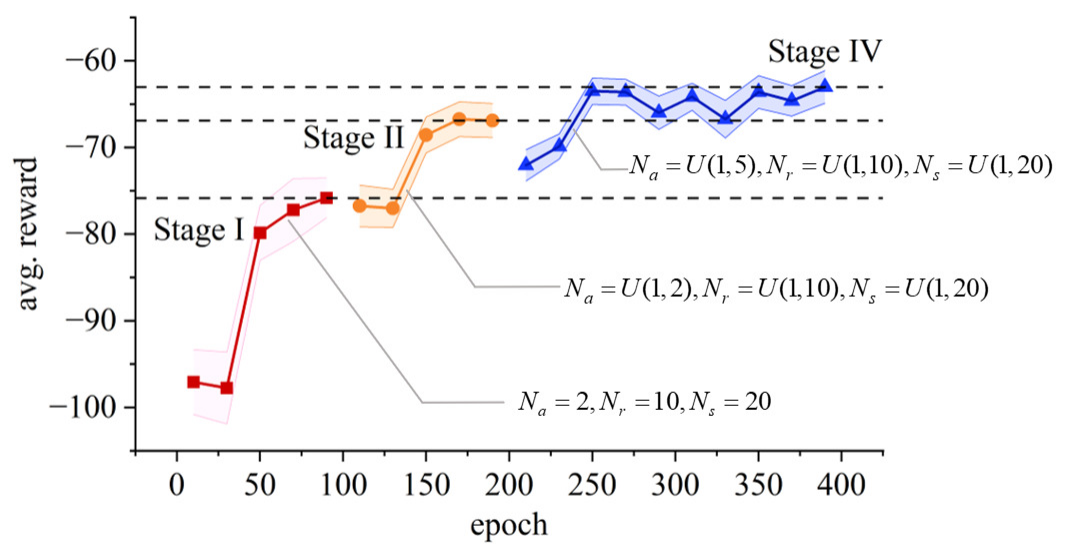}
	\caption{The curriculum learning results for three stages.}
	\label{course learning}
\end{figure}
Fig. \ref{course learning} depicts the training curves of the average reward for HCR2C during the training process of curriculum learning. The reward values at all stages are uniformly validated on 1024 randomly generated instances at U3 scale on M3. The training process is divided into three stages: in stage \uppercase\expandafter{\romannumeral1}, HCR2C is trained on F2 fixed-scale instances. It can be observed from the red curve that HCR2C demonstrates a certain level of generalization ability in the first stage; however, due to the limited training instance scale, the reward value reaches a bottleneck and exhibits a larger variance. In stage \uppercase\expandafter{\romannumeral2}, we randomize the number of robots, the number of retrieval racks and unoccupied storage locations in a small range. In stage \uppercase\expandafter{\romannumeral3}, we further expand the scale of randomization range: the instances are randomly generated with a uniform distribution $U(1,5)$ for robots, $U(1,10)$ for retrieval racks, and $U(1,20)$ for unoccupied storage locations. 
Upon validation of the learning curve at various stages, it is apparent that each stage demonstrates a fluctuating upward trajectory. The overall reward trend in Fig. \ref{course learning} continues to ascend although the  randomness and difficulty of curricula are gradually increasing, indicating HCR2C's ongoing enhancement in generalizing instances of random scales and the experience learned from the lower stage curriculum are not forgotten. The increase in reward for policies within a higher stage is not as significant as that in lower stage policies, suggesting that the learning process tends towards stability and convergence during curriculum learning.

\subsubsection{Generalization Performance Analysis}

\begin{figure*}[!t]
	\centering
	\includegraphics[width=5.15in]{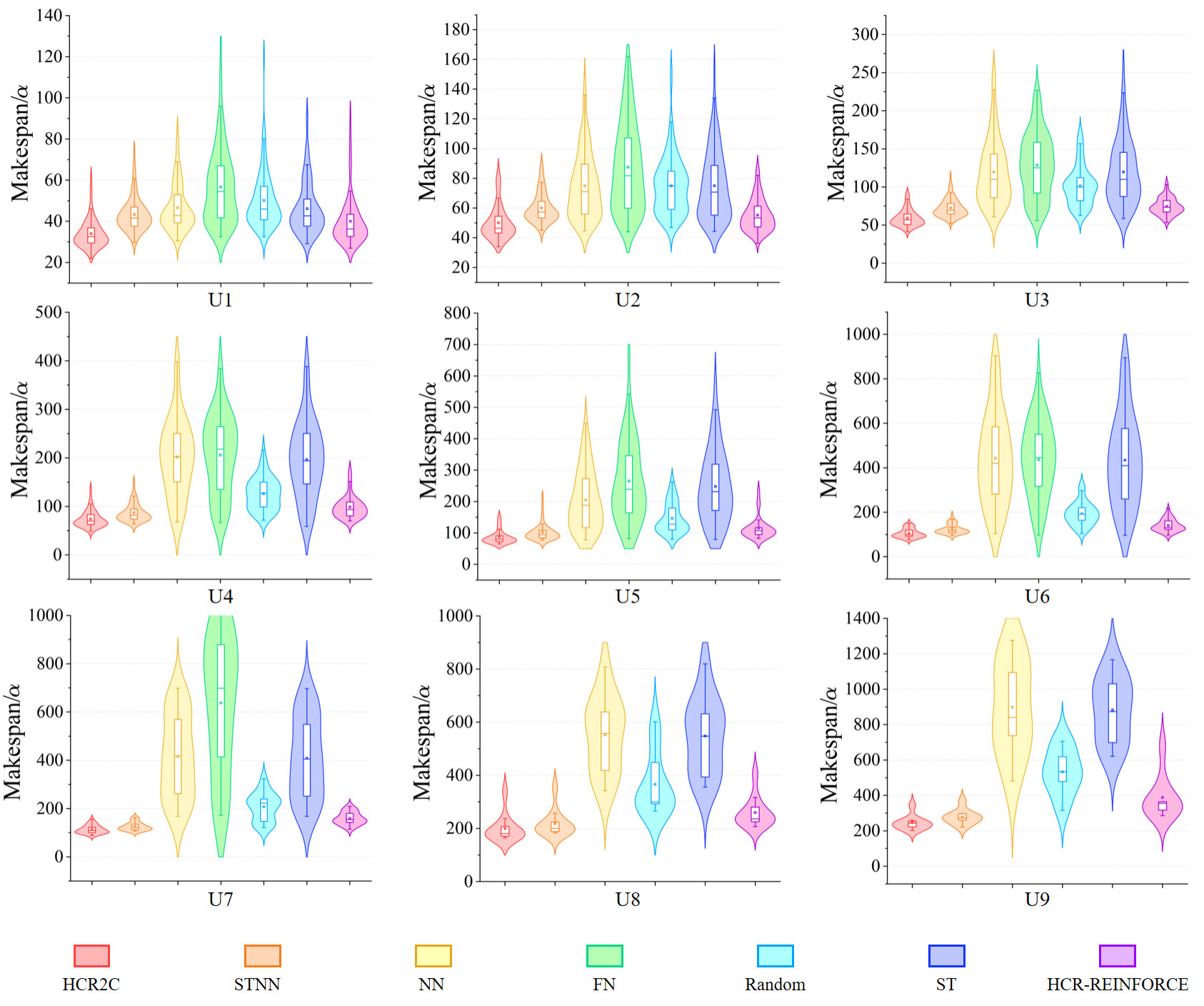}
	\caption{Violin plot of test distribution results of quality performance with different methods on U1-U9 random scale simulated instances.}
	\label{randomgraph}
\end{figure*}

\begin{figure}[!t]
	\centering
	\includegraphics[scale=0.66]{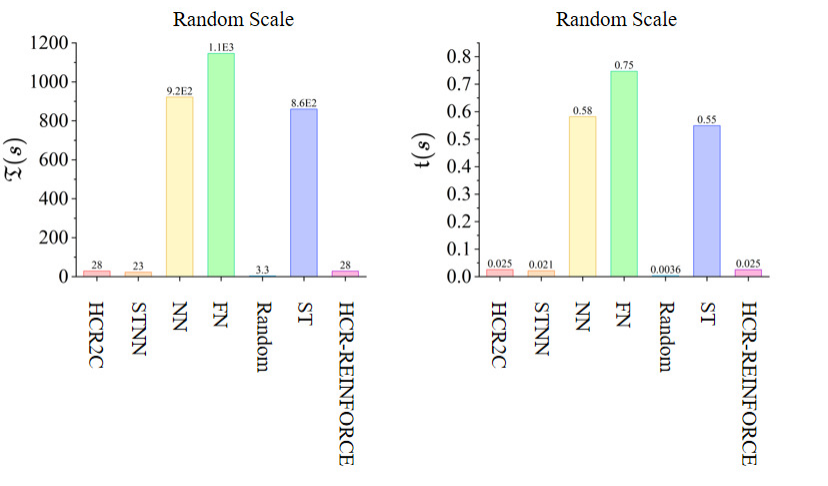}
	\caption{Comparison results of running time indicator with different methods on random scale simulated instances.}
	\label{wholet2}
\end{figure}

\begin{table*}[!t]
	\caption{RESULTS OF QUALITY PERFORMANCE ON U1-U9 RANDOM SCALE SIMULATED INSTANCES.
		\label{tab:randomtest}}
	%\belowrulesep=2pt
	%\aboverulesep=2pt
	\renewcommand{\arraystretch}{1.1}	
	\centering
	\resizebox{18.2cm}{!}{
		\begin{tabular}{c c cc cc cc cc cc cc}
			%\hline 
			\Xhline{1pt}
			%\multicolumn{11}{|l|} \\
			%\hline \hline
			%\multirow{2}{*}{}
			{Instance}
			& \multicolumn{1}{c}{{HCR2C}} &  \multicolumn{2}{c}{{HCR-REINFORCE}}  & \multicolumn{2}{c}{{STNN}} &  \multicolumn{2}{c}{{ST}} & \multicolumn{2}{c}{{NN}} & \multicolumn{2}{c}{{FN}}  & \multicolumn{2}{c}{{Random}}    \\
			%\cline{3-4} \cline{4-5}
			\cmidrule(lr){2-2} 
			\cmidrule(lr){3-4} \cmidrule(lr){5-6}  \cmidrule(lr){7-8}  \cmidrule(lr){9-10} \cmidrule(lr){11-12} \cmidrule(lr){13-14} 
			{ID}&$\mathfrak{W}(s)$ &$\mathfrak{W}(s)$ &Gap & $\mathfrak{W}(s)$ &Gap & $\mathfrak{W}(s)$ &Gap & $\mathfrak{W}(s)$ &Gap & $\mathfrak{W}(s)$ &Gap & $\mathfrak{W}(s)$ &Gap   \\
			\Xhline{0.7pt}
			U1&	{33.98} &	40.01& 	17.77\%& 	43.31& 	27.47\% &	46.16& 	   35.85\% &	46.60& 	37.15\%&	56.70& 	66.89\%&	50.11& 	47.48\% \\
			U2&	{50.03} &	55.18& 	10.29\%& 	60.15& 	20.21\% &	75.00& 	    49.90\% &	74.88& 	49.65\%&	87.51& 	74.90\%&	74.93& 	49.75\%   \\
			U3&	{59.35} &	74.52& 	25.55\%& 	72.43& 	22.04\% &	119.73& 	101.72\%& 	119.70& 101.67\%&	128.54& 	116.57\%&	101.55& 	71.09\%   \\
			U4&	{73.85} &	98.44& 	33.31\% & 	86.76& 	17.49\%&	201.81& 	173.29\%&	196.50& 166.10\%&	205.76& 	178.63\%&	126.41& 	71.19\%  \\
			U5&	{86.33} &	115.09& 33.32\%& 	101.54& 17.62\%&	205.21&     137.72\%&	247.73& 186.98\%&	265.26& 	207.28\%&	146.50& 	69.70\%   \\
			U6&	{105.37} & 140.22& 33.07\%& 	123.92& 17.60\%&	442.63&     320.05\% &	434.35& 312.19\%&	436.55& 	314.28\%&	193.51& 	83.64\%\\
			U7&	{115.28} &160.38& 39.11\%& 	126.90& 10.08\%&	415.54& 	260.45\% &	406.95& 253.00\%&	631.76& 	448.01\%&	207.22& 	79.75\%   \\
			U8&	{199.13} &268.02& 34.60\%& 	218.03&  9.49\%&	551.71& 	177.06\% &	547.76& 175.07\%&	2.01E3& 	907.02\%&	365.82& 	83.71\% \\
			U9&	{244.81} &387.66& 58.35\%& 	276.06& 12.76\%&	897.53 & 	266.62\% &	882.73&	260.57\%&	4.64E3& 	1.80E3\%&	579.92& 	136.88\%\\
			%\hline
			\Xhline{0.7pt}
			\textbf{AVG.} &	    &		&31.71\% &	      &{17.20\%} & 		&171.26\%  &     &169.30\%	&   	&456.70\%	&		&	77.02\%   \\
			\Xhline{1pt}
		\end{tabular}
	}
\end{table*}

To test the performance of scaling up and generalization of the universal policy after trained with HCR2C,
we conduct a comparison experiment between HCR2C, HCR-REINFORCE and heuristic methods on 100 random simulated instances with the scale configuration scheme of U1-U9, respectively, where maps have never been trained on except for M3. Two new heuristic methods are added in this comparison experiments because heuristic methods have natural advantages on random-scaled MRTP.  The new heuristic methods are adjusted to obey our framework, where Farthest Neighbor (FN) \cite{bibtex6} chooses the valid pair of graph node location and corresponding robot with the largest evaluation traveling time between current robot locations and their farthest node locations as the planning result for next step.
Shortest Time (ST) chooses the valid pair of location node and corresponding robot with the smallest evaluated traveling time considering history robot traveling time and time consuming between current robot locations and their nearest node locations as the next step planning result. 

On random scale simulated instances,
it is evident that the random variation of scale has a significant impact on the makespan's indicator results. In order to better analyze quality performance of methods, we design a new quality indicator $\mathfrak{W}=\mathfrak{V}/\alpha$ to evaluate the generalization quality, where $\alpha=N_r/{N_a}$ is designed to mitigate issues arising from large variance in makespan caused by various instance scale.
Table \ref{tab:randomtest} presents the $\mathfrak{W}$ indicator gaps among different methods and HCR2C. It is evident that HCR2C performs best among them, and the second-best method STNN still exhibits an average gap of $17.20\%$ compared to HCR2C. Test results on the random scale simulated instances reveal a $31.71\%$ gap between HCR2C and HCR-REINFORCE, indicating the significant effect of curriculum learning. The result of HCR-REINFORCE demonstrates poor generalization, with its  $\mathfrak{W}$ being inferior to that of heuristic method STNN.
To further analyze the stability of algorithm performance, we present the data distribution and important statistics of $\mathfrak{W}$ indicator using a violin graph. Fig. \ref{randomgraph} illustrates that HCR2C's $\mathfrak{W}$ values remain relatively stable across U1-U9 random scale simulated instances on various maps including that have never been learned in training process. In contrast, NN, FN, and random policy exhibit significant deterioration in stability as the scale increases. Specifically, on a larger scale, the mean value of $\mathfrak{W}$ in Random algorithm is lower than that of NN and FN algorithms, consistent with observations in fixed-scale simulated instances. This suggests that these methods face greater challenges with an increasing scale. It is noteworthy that after the robot scale exceeds 100, HCR-REINFORCE's $\mathfrak{W}$ data distribution becomes dispersed. However, HCR2C overcomes these issues after a few training of curriculum learning.
Fig. \ref{wholet2} presents the running time for various methods in solving MRTP in RMFS on different random scales, including the generation time of a single joint option. Although HCR-REINFORCE shows slower speed compared to STNN and Random, it is still capable of generating step-wise robot task planning instructions within $0.1s$ for hyper and random scale instances involving up to 200 robots, enabling real-time planning at the millisecond level for multi-robot tasks on a hyper scale MRTP in RMFS.

\subsection{Experiments On Real-World Instances}
\subsubsection{Real-World Setup}
\begin{figure}[!t]
	\centering
	\includegraphics[width=3.1in]{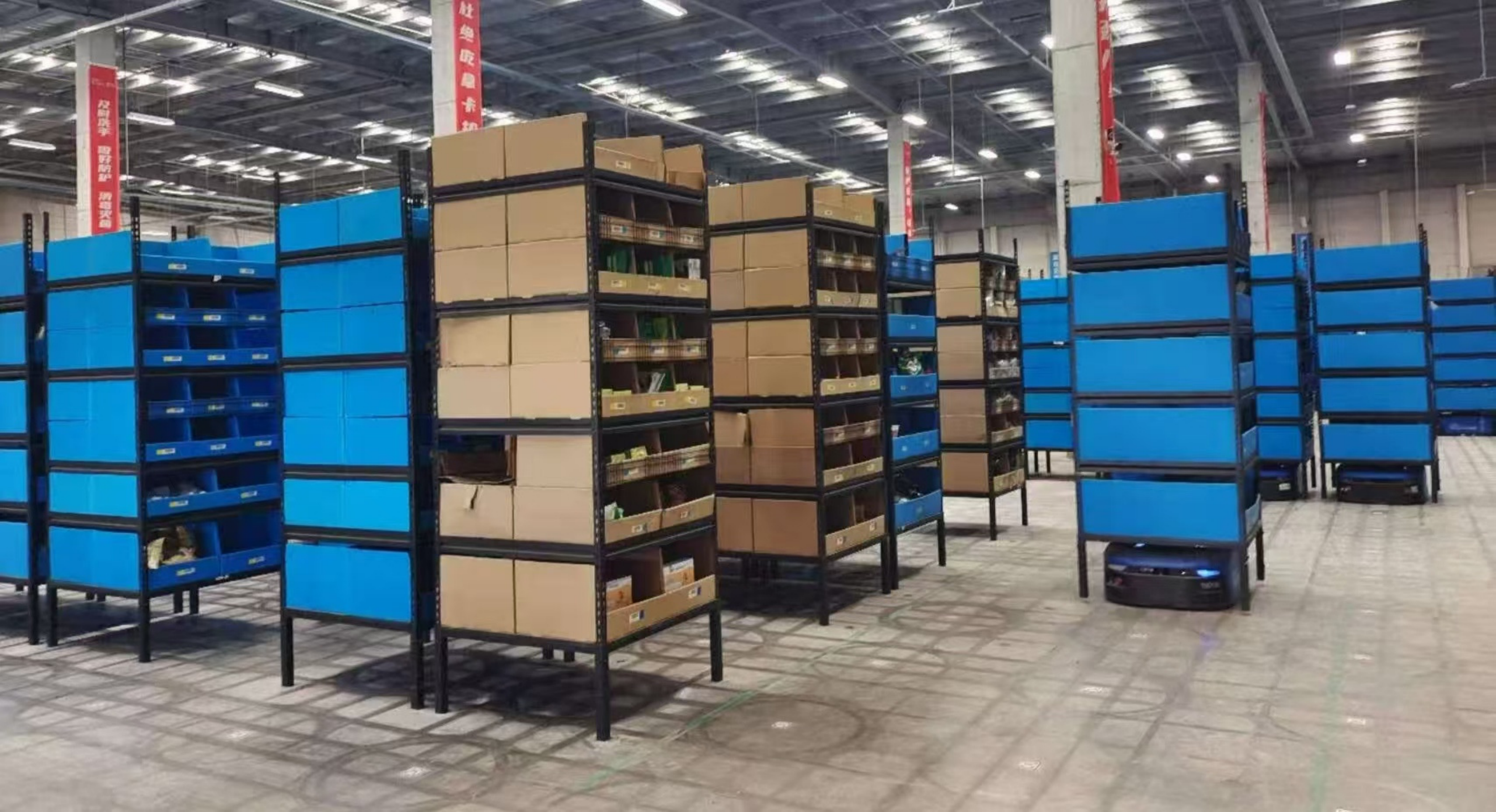}
	\caption{The operation scene in real-world RMFS.}
	\label{rmfs}
\end{figure}
To further evaluate the generalization and practical performance of our planner, we prepare real-world business instances for tests. These instances are obtained from a real-world RMFS with a cargo-to-person model situated constructed by the most famous warehousing and logistics company in China. The operation scene of this real-world RMFS is depicted in Fig. \ref{rmfs}, encompassing a total of 20 picking stations, 264 robots, and 2164 racks and storage locations.  The average moving speed for mobile robots is $2m/s$ without rack and $1.6m/s$ with rack. Considering the speed factors in obstacles, turning, and pausing, we estimate the average moving speed of the mobile robot  as $v_r=1m/s$. 
%The map layout is never learned before, and the maximum valid locations in each zone is not fixed.

The real-world instances are obtained after the RMFS system receives the top-level order demands and the orders are processed by the order allocation system. This data encompasses information such as the index, current location, target location, current moving speed, power consumption, and other status details of mobile robots. It also includes the index of retrieval racks to be moved, along with their respective locations. Additionally, it covers occupation status of each storage location, index of picking stations for retrieval racks to be sent to, picking station locations, and global task planning information such as clock time. From these data, we randomly selected 20 sets of real-world business data as our real-world instances, and each instance is associated with a multi-robot system's task planning cycle where robots are required to move all retrieval racks to their assigned picking stations.

\subsubsection{Test Results and Performance Analysis}

\begin{figure}[!t]
	\centering
	\includegraphics[scale=0.37]{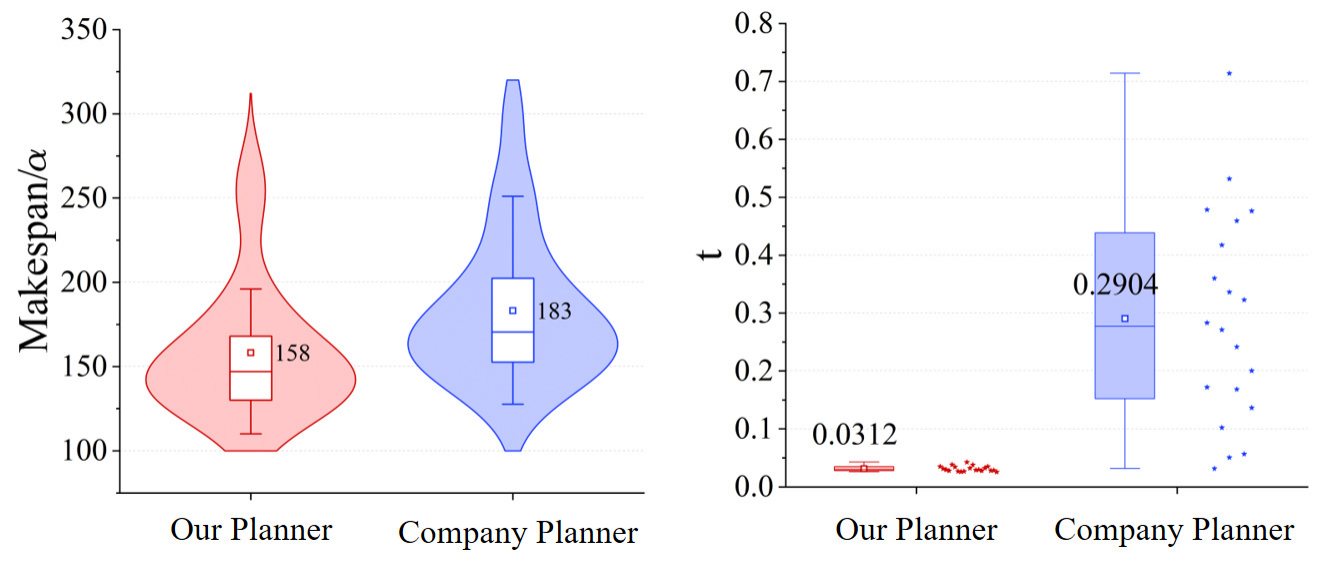}
	\caption{Comparison results of our planner and native company method on real-world RMFS instances.}
	\label{realwt}
\end{figure}

We compare our planner generalized with HCR2C against the native company task planner in 20 sets of real-world instances. Statistical results of the comparison experiment are presented in Fig. \ref{realwt}. It is obvious that our planner outperforms the company's method with a higher planning quality and faster planning speed to generate multi-robot task planning scheme. Our planner obtains an average quality indicator $\mathfrak{W} = 158.2s$, lower than the company's $183.14s$. 
Also, both in the best and worst cases, the quality performance of our planner is better. 
On average, our planner reduces $\mathfrak{W}$ performance by $13.62\%$ compared to the company's method, and the average task planning instruction generation time is $0.031s$ for our method, much lower than the company's $0.29s$. 
The planning speed performance of our planner is very stable across different instances, as is also evident from
the distribution of one step planning time $\mathfrak{t}$  in Fig. \ref{realwt}.
Thus, our planner can be successfully scaled up to real-world RMFS instances that have never been learned before and achieve faster and better performance than native company planners.

\section{Conclusion}
\label{sec:conclusion}
In this work, we constructed a scalable multi-stage HRL-based multi-robot task planner for hyper scale MRTP in RMFS that shows challenges in dimension disaster, dynamic properties, scaling up and generalization. 
A feasible planning scheme for solving these challenges based on the centralized architecture to ensure optimality was presented.
For modeling, we proposed a universal concept of C2AMRTG whose graph topology is unfixed and differs between robots and global system. 
Based on the C2AMRTG, we used the MDP with options on C2AMRTG to model the problem, and the DRL framework provides a natural dynamic planning implementation.
The C2AMRTG also abstracted the key and understandable features for MRTP in RMFS and gave us inspirations to design network. The ablation study showed our HTAN with temporal embedding layers does not depend on all historical state information but can achieve better performance.
For training, we noted that HRL has credit assignment challenges, and proposed a new HRL algorithm HCR-REINFORCE to improve training efficiency for the integrated multi-layer policies. 
According to the results of experiments, HCR-REINFORCE showed faster convergence and higher rewards compared with other learning methods. It also obtained higher quality planning performance compared with heuristic methods, but its performance degraded on unlearned scales and maps. To tackle this problem, we designed HCR2C to teach policies how to scale up and generalize well. 
Experimental results demonstrated that our planner with these methods performed best among different kinds of state-of-the-art methods while successfully scaling up to random instances with up to 200 robots on various unlearned maps and outperformed native company planner on real-world instances.
Based on this research, our future work will continue to expand the current MRTP with more policy layers and heterogeneous robots. Additionally, we will investigate the automatic hierarchical decomposition technology for intelligently addressing tasks with more intricate structures.

\bibliographystyle{IEEEtran}
\bibliography{bibtex/BIB-mypaper}

\begin{thebibliography}{10}
\providecommand{\url}[1]{#1}
\csname url@samestyle\endcsname
\providecommand{\newblock}{\relax}
\providecommand{\bibinfo}[2]{#2}
\providecommand{\BIBentrySTDinterwordspacing}{\spaceskip=0pt\relax}
\providecommand{\BIBentryALTinterwordstretchfactor}{4}
\providecommand{\BIBentryALTinterwordspacing}{\spaceskip=\fontdimen2\font plus
\BIBentryALTinterwordstretchfactor\fontdimen3\font minus
  \fontdimen4\font\relax}
\providecommand{\BIBforeignlanguage}[2]{{%
\expandafter\ifx\csname l@#1\endcsname\relax
\typeout{** WARNING: IEEEtran.bst: No hyphenation pattern has been}%
\typeout{** loaded for the language `#1'. Using the pattern for}%
\typeout{** the default language instead.}%
\else
\language=\csname l@#1\endcsname
\fi
#2}}
\providecommand{\BIBdecl}{\relax}
\BIBdecl

\bibitem{trends}
H.~Guo, F.~Wu, Y.~Qin, R.~Li, K.~Li, and K.~Li, ``Recent trends in task and
  motion planning for robotics: A survey,'' \emph{ACM Computing Surveys},
  vol.~55, no. 13s, pp. 1--36, 2023.

\bibitem{MRTP}
L.~Antonyshyn, J.~Silveira, S.~Givigi, and J.~Marshall, ``Multiple mobile robot
  task and motion planning: A survey,'' \emph{ACM Computing Surveys}, vol.~55,
  no.~10, pp. 1--35, 2023.

\bibitem{schedule}
A.~Ham, ``Transfer-robot task scheduling in job shop,'' \emph{International
  Journal of Production Research}, vol.~59, no.~3, pp. 813--823, 2021.

\bibitem{allocation}
G.~A. Korsah, A.~Stentz, and M.~B. Dias, ``A comprehensive taxonomy for
  multi-robot task allocation,'' \emph{The International Journal of Robotics
  Research}, vol.~32, no.~12, pp. 1495--1512, 2013.

\bibitem{decompose}
M.~Guo and D.~V. Dimarogonas, ``Task and motion coordination for heterogeneous
  multiagent systems with loosely coupled local tasks,'' \emph{IEEE
  Transactions on Automation Science and Engineering}, vol.~14, no.~2, pp.
  797--808, 2016.

\bibitem{RMFS}
A.~Rim{\'e}l{\'e}, M.~Gamache, M.~Gendreau, P.~Grangier, and L.-M. Rousseau,
  ``Robotic mobile fulfillment systems: a mathematical modelling framework for
  e-commerce applications,'' \emph{International Journal of Production
  Research}, pp. 1--17, 2021.

\bibitem{Warehouse}
N.~Boysen, R.~De~Koster, and F.~Weidinger, ``Warehousing in the e-commerce era:
  A survey,'' \emph{European Journal of Operational Research}, vol. 277, no.~2,
  pp. 396--411, 2019.

\bibitem{bibtex1}
F.~B. Sorbelli, S.~Carpin, F.~Coro, S.~K. Das, A.~Navarra, and C.~M. Pinotti,
  ``Speeding up routing schedules on aisle graphs with single access,''
  \emph{IEEE Transactions on Robotics}, vol.~38, no.~1, pp. 433--447, 2021.

\bibitem{camisa2022multi}
A.~Camisa, A.~Testa, and G.~Notarstefano, ``Multi-robot pickup and delivery via
  distributed resource allocation,'' \emph{IEEE Transactions on Robotics},
  vol.~39, no.~2, pp. 1106--1118, 2022.

\bibitem{bibtex3}
Y.~Zhuang, Y.~Zhou, E.~Hassini, Y.~Yuan, and X.~Hu, ``Rack retrieval and
  repositioning optimization problem in robotic mobile fulfillment systems,''
  \emph{Transportation Research Part E: Logistics and Transportation Review},
  vol. 167, p. 102920, 2022.

\bibitem{shi2023bi}
X.~Shi, F.~Deng, S.~Lu, Y.~Fan, L.~Ma, and J.~Chen, ``A bi-level optimization
  approach for joint rack sequencing and storage assignment in robotic mobile
  fulfillment systems,'' \emph{Science China Information Sciences}, vol.~66,
  no.~11, p. 212202, 2023.

\bibitem{bibtex6}
A.~Gharehgozli and N.~Zaerpour, ``Robot scheduling for pod retrieval in a
  robotic mobile fulfillment system,'' \emph{Transportation Research Part E:
  Logistics and Transportation Review}, vol. 142, p. 102087, 2020.

\bibitem{ha2021warehouse}
W.~Y. Ha, L.~Cui, and Z.-P. Jiang, ``A warehouse scheduling using genetic
  algorithm and collision index,'' in \emph{2021 20th International Conference
  on Advanced Robotics (ICAR)}.\hskip 1em plus 0.5em minus 0.4em\relax IEEE,
  2021, pp. 318--323.

\bibitem{bibtex5}
T.~Ren, J.~Niu, X.~Liu, J.~Wu, X.~Lei, and Z.~Zhang, ``An efficient model-free
  approach for controlling large-scale canals via hierarchical reinforcement
  learning,'' \emph{IEEE Transactions on Industrial Informatics}, vol.~17,
  no.~6, pp. 4367--4378, 2020.

\bibitem{ma2024efficient}
C.~Ma, A.~Li, Y.~Du, H.~Dong, and Y.~Yang, ``Efficient and scalable
  reinforcement learning for large-scale network control,'' \emph{Nature
  Machine Intelligence}, pp. 1--15, 2024.

\bibitem{mazyavkina2021reinforcement}
N.~Mazyavkina, S.~Sviridov, S.~Ivanov, and E.~Burnaev, ``Reinforcement learning
  for combinatorial optimization: A survey,'' \emph{Computers \& Operations
  Research}, vol. 134, p. 105400, 2021.

\bibitem{pateria2021hierarchical}
S.~Pateria, B.~Subagdja, A.-h. Tan, and C.~Quek, ``Hierarchical reinforcement
  learning: A comprehensive survey,'' \emph{ACM Computing Surveys (CSUR)},
  vol.~54, no.~5, pp. 1--35, 2021.

\bibitem{vinyals2019grandmaster}
O.~Vinyals, I.~Babuschkin, W.~M. Czarnecki, M.~Mathieu, A.~Dudzik, J.~Chung,
  D.~H. Choi, R.~Powell, T.~Ewalds, P.~Georgiev \emph{et~al.}, ``Grandmaster
  level in starcraft ii using multi-agent reinforcement learning,''
  \emph{Nature}, vol. 575, no. 7782, pp. 350--354, 2019.

\bibitem{lee2024learning}
J.~Lee, M.~Bjelonic, A.~Reske, L.~Wellhausen, T.~Miki, and M.~Hutter,
  ``Learning robust autonomous navigation and locomotion for wheeled-legged
  robots,'' \emph{Science Robotics}, vol.~9, no.~89, p. eadi9641, 2024.

\bibitem{2021scheduling}
V.~Tereshchuk, N.~Bykov, S.~Pedigo, S.~Devasia, and A.~G. Banerjee, ``A
  scheduling method for multi-robot assembly of aircraft structures with soft
  task precedence constraints,'' \emph{Robotics and Computer-Integrated
  Manufacturing}, vol.~71, p. 102154, 2021.

\bibitem{samiei2023distributed}
A.~Samiei and L.~Sun, ``Distributed matching-by-clone hungarian-based algorithm
  for task allocation of multi-agent systems,'' \emph{IEEE Transactions on
  Robotics}, 2023.

\bibitem{motes2020multi}
J.~Motes, R.~Sandstr{\"o}m, H.~Lee, S.~Thomas, and N.~M. Amato, ``Multi-robot
  task and motion planning with subtask dependencies,'' \emph{IEEE Robotics and
  Automation Letters}, vol.~5, no.~2, pp. 3338--3345, 2020.

\bibitem{fu2022robust}
B.~Fu, W.~Smith, D.~M. Rizzo, M.~Castanier, M.~Ghaffari, and K.~Barton,
  ``Robust task scheduling for heterogeneous robot teams under capability
  uncertainty,'' \emph{IEEE Transactions on Robotics}, vol.~39, no.~2, pp.
  1087--1105, 2022.

\bibitem{luo2022temporal}
X.~Luo and M.~M. Zavlanos, ``Temporal logic task allocation in heterogeneous
  multirobot systems,'' \emph{IEEE Transactions on Robotics}, vol.~38, no.~6,
  pp. 3602--3621, 2022.

\bibitem{enright2011optimization}
J.~J. Enright and P.~R. Wurman, ``Optimization and coordinated autonomy in
  mobile fulfillment systems,'' in \emph{Workshops at the twenty-fifth AAAI
  conference on artificial intelligence}, 2011.

\bibitem{MATSPsolve}
S.~C. Sarin, H.~D. Sherali, J.~D. Judd, and P.-F.~J. Tsai, ``Multiple
  asymmetric traveling salesmen problem with and without precedence
  constraints: Performance comparison of alternative formulations,''
  \emph{Computers \& Operations Research}, vol.~51, pp. 64--89, 2014.

\bibitem{dynamic4}
D.~Shi, Y.~Tong, Z.~Zhou, K.~Xu, W.~Tan, and H.~Li, ``Adaptive task planning
  for large-scale robotized warehouses,'' in \emph{2022 IEEE 38th International
  Conference on Data Engineering (ICDE)}.\hskip 1em plus 0.5em minus
  0.4em\relax IEEE, 2022, pp. 3327--3339.

\bibitem{cheng2024deep}
B.~Cheng, T.~Xie, L.~Wang, Q.~Tan, and X.~Cao, ``Deep reinforcement learning
  driven cost minimization for batch order scheduling in robotic mobile
  fulfillment systems,'' \emph{Expert Systems with Applications}, vol. 255, p.
  124589, 2024.

\bibitem{spatio}
Y.~Lian, Q.~Yang, Y.~Liu, and W.~Xie, ``A spatio-temporal constrained
  hierarchical scheduling strategy for multiple warehouse mobile robots under
  industrial cyber--physical system,'' \emph{Advanced Engineering Informatics},
  vol.~52, p. 101572, 2022.

\bibitem{kleinert2021survey}
T.~Kleinert, M.~Labb{\'e}, I.~Ljubi{\'c}, and M.~Schmidt, ``A survey on
  mixed-integer programming techniques in bilevel optimization,'' \emph{EURO
  Journal on Computational Optimization}, vol.~9, p. 100007, 2021.

\bibitem{kaufmann2023champion}
E.~Kaufmann, L.~Bauersfeld, A.~Loquercio, M.~M{\"u}ller, V.~Koltun, and
  D.~Scaramuzza, ``Champion-level drone racing using deep reinforcement
  learning,'' \emph{Nature}, vol. 620, no. 7976, pp. 982--987, 2023.

\bibitem{koolattention}
W.~Kool, H.~van Hoof, and M.~Welling, ``Attention, learn to solve routing
  problems!'' in \emph{International Conference on Learning Representations},
  2019.

\bibitem{zhang2020learning}
C.~Zhang, W.~Song, Z.~Cao, J.~Zhang, P.~S. Tan, and X.~Chi, ``Learning to
  dispatch for job shop scheduling via deep reinforcement learning,''
  \emph{Advances in neural information processing systems}, vol.~33, pp.
  1621--1632, 2020.

\bibitem{a2c}
V.~Mnih, ``Asynchronous methods for deep reinforcement learning,'' \emph{arXiv
  preprint arXiv:1602.01783}, 2016.

\bibitem{REINFORCE}
R.~J. Williams, ``Simple statistical gradient-following algorithms for
  connectionist reinforcement learning,'' \emph{Machine learning}, vol.~8, pp.
  229--256, 1992.

\bibitem{JPPO}
I.~Radosavovic, T.~Xiao, B.~Zhang, T.~Darrell, J.~Malik, and K.~Sreenath,
  ``Real-world humanoid locomotion with reinforcement learning,'' \emph{Science
  Robotics}, vol.~9, no.~89, p. eadi9579, 2024.

\bibitem{PPO}
J.~Schulman, F.~Wolski, P.~Dhariwal, A.~Radford, and O.~Klimov, ``Proximal
  policy optimization algorithms,'' \emph{arXiv preprint arXiv:1707.06347},
  2017.

\bibitem{ho2022federated}
T.~M. Ho, K.-K. Nguyen, and M.~Cheriet, ``Federated deep reinforcement learning
  for task scheduling in heterogeneous autonomous robotic system,'' \emph{IEEE
  Transactions on Automation Science and Engineering}, vol.~21, no.~1, pp.
  528--540, 2022.

\bibitem{kostrikin1982introduction}
A.~I. Kostrikin and R.~A. Sala, \emph{Introduction to algebra}.\hskip 1em plus
  0.5em minus 0.4em\relax Springer, 1982, vol.~8.

\bibitem{sutton1999between}
R.~S. Sutton, D.~Precup, and S.~Singh, ``Between mdps and semi-mdps: A
  framework for temporal abstraction in reinforcement learning,''
  \emph{Artificial intelligence}, vol. 112, no. 1-2, pp. 181--211, 1999.

\bibitem{vaswani2017attention}
A.~Vaswani, N.~Shazeer, N.~Parmar, J.~Uszkoreit, L.~Jones, A.~N. Gomez,
  {\L}.~Kaiser, and I.~Polosukhin, ``Attention is all you need,''
  \emph{Advances in neural information processing systems}, vol.~30, 2017.

\bibitem{sutton2018reinforcement}
R.~S. Sutton, ``Reinforcement learning: An introduction,'' \emph{A Bradford
  Book}, pp. 325--326, 2018.

\bibitem{foerster2018counterfactual}
J.~Foerster, G.~Farquhar, T.~Afouras, N.~Nardelli, and S.~Whiteson,
  ``Counterfactual multi-agent policy gradients,'' in \emph{Proceedings of the
  AAAI conference on artificial intelligence}, vol.~32, no.~1, 2018.

\bibitem{BC}
F.~Torabi, G.~Warnell, and P.~Stone, ``Behavioral cloning from observation,''
  \emph{arXiv preprint arXiv:1805.01954}, 2018.

\bibitem{NN}
F.~Weidinger and N.~Boysen, ``Scattered storage: How to distribute stock
  keeping units all around a mixed-shelves warehouse,'' \emph{Transportation
  Science}, vol.~52, no.~6, pp. 1412--1427, 2018.

\end{thebibliography}

\end{document}